\UseRawInputEncoding
\documentclass{article}
\usepackage[utf8]{inputenc}
\usepackage{arxiv}
\usepackage{url} 
\usepackage{lineno}
\usepackage{soul}

\usepackage{bm}
\usepackage{amsmath}
\usepackage{amsthm}
\usepackage{amssymb}
\usepackage{graphicx}
\usepackage{subcaption}
\usepackage{bbold}

\usepackage{listings} 

\usepackage{algorithm}
\usepackage{algpseudocode}

\usepackage[square,numbers]{natbib}

\usepackage[parfill]{parskip}

\usepackage{enumitem}

\usepackage[final]{hyperref}
\usepackage[nameinlink]{cleveref}

\newcommand{\nModel}{m} 
\newcommand{\state}{\bm{\psi}}
\newcommand{\trueState}{\state^{*}}
\newcommand{\stateSpace}{\mathbb{R}^{\nModel}}
\newcommand{\timeIndex}{t}
\newcommand{\dynamicFwd}{\mathcal{F}}

\newcommand{\modelNoise}{\bm{\delta}}
\newcommand{\modelNoiseCov}{\bm{\Delta}}

\newcommand{\fwd}{\bm{G}}
\newcommand{\nData}{n}

\newcommand{\dataVec}{\bm{y}}
\newcommand{\noise}{\bm{\epsilon}}
\newcommand{\noiseCov}{\bm{E}}

\newcommand{\randomState}{\bm{\Psi}} 
\newcommand{\mean}{\bm{\mu}}
\newcommand{\cov}{\bm{\Sigma}}
\newcommand{\kalmanGain}{\bm{K}}

\newcommand{\condMean}{\bm{\mu}}
\newcommand{\condCov}{\bm{\Sigma}}

\newcommand{\forecastMean}{\bm{\mu^f}}
\newcommand{\forecastCov}{\bm{\Sigma^f}}

\newcommand{\ensMember}{\state}

\newcommand{\ensMemberI}{\ensMember^{(i)}}
\newcommand{\ensMemberIf}{\ensMember^{f(i)}} 
\newcommand{\ensMemberJf}{\ensMember^{f(j)}} 
\newcommand{\nEns}{p}

\newcommand{\kalmanGainEns}{\bm{\hat{K}}}
\newcommand{\estForecastCov}{\bm{\hat{\Sigma}^f}} 

\newcommand{\ensMean}{\bar{\state}}
\newcommand{\ensDevI}{\ensMember^{(i)'}}
\newcommand{\ensDevIf}{\ensMember^{f(i)'}}
\newcommand{\ensMeanForecast}{\bar{\state}^f}
\newcommand{\kalmanGainSR}{\bm{\tilde{K}}}

\newcommand{\sampleCov}{\hat{S}}

\newcommand{\locMatrix}{\bm{\rho}}
\newcommand{\locFun}{\rho}

\newcommand{\spaceDim}{d} 

\newcommand{\nGrid}{\nModel}

\newcommand{\estCov}{\bm{\hat{\Sigma}}} 
\newcommand{\singVecL}{u}
\newcommand{\singVal}{\lambda}
\newcommand{\svdRank}{k} 

\newcommand{\noiseStd}{\sigma_{\epsilon}}

\newcommand{\stateInd}{\ell} 

\newcommand{\dataIndex}{j} 

\title{Non-Sequential Ensemble Kalman Filtering using Distributed Arrays}

\usepackage{authblk}

\author[1,2,4]{%
	C\'edric Travelletti\thanks{\texttt{cedric.travelletti@unibe.ch}}%
}
\author[3,4]{%
	J\"org Franke\thanks{\texttt{joerg.franke@unibe.ch}}%
}
\author[2,3]{%
	David Ginsbourger%
}
\author[3,4]{%
	Stefan Br\"onnimann%
}
\affil[1]{Mathematics for Materials Modelling, Institute of Mathematics \& Institute of Materials, EPFL, Lausanne, Switzerland}
\affil[2]{Institute of Mathematical Statistics and Actuarial Science, University of Bern, Bern, Switzerland}
\affil[3]{Oeschger Center for Climate Change Research, University of Bern, Bern, Switzerland}
\affil[4]{Institute of Geography, University of Bern, Bern, Switzerland}

\renewcommand{\headeright}{\phantom{dafas}}
\renewcommand{\undertitle}{\phantom{dafas}}
\renewcommand{\shorttitle}{\phantom{dafas}}

\hypersetup{
pdftitle={Non-Sequential Ensemble Kalman Filtering using Distributed Arrays},
pdfsubject={stat.ML},
pdfauthor={C\'edric Travelletti, J\"org Franke, David Ginsbourger, Stefan Br\"onnimann},
pdfkeywords={ensemble Kalman filtering, data assimilation, distributed computing},
}

\begin{document}

\maketitle

\begin{abstract}
This work introduces a new, distributed implementation of the Ensemble Kalman Filter (EnKF) 
that allows for non-sequential assimilation of large datasets in high-dimensional problems.
The traditional EnKF  algorithm is computationally intensive and exhibits difficulties in applications requiring interaction 
with the background covariance matrix, prompting the use of methods like sequential assimilation which can introduce unwanted consequences, 
such as dependency on observation ordering. 
Our implementation leverages recent advancements in distributed computing to enable the construction and use of the full model error covariance matrix in distributed memory, 
allowing for single-batch assimilation of all observations and eliminating order dependencies. 
Comparative performance assessments, involving both synthetic and real-world paleoclimatic reconstruction applications, 
indicate that the new, non-sequential implementation outperforms the traditional, sequential one.

\end{abstract}
\newpage

\section{Introduction}\label{sec:introduction}
Kalman filtering is a data assimilation method that aims at estimating 
the state of a dynamic system from a series of measurements. 
It has been used successfully in a variety of domains such as 
aerospace engineering, robotics, navigation systems, financial economics and climate science.
In spite of these achievements, the original Kalman filter suffers from several limitations 
that prevent its application at larger scales. This has given rise to a variety of 
modified Kalman filters, such as the Extended Kalman Filter (EKF) 
\citep{gelb1974applied,lefebvre2004kalman}, 
the Unscented Kalman Filter (UKF) \citep{julier1997new,wan2000unscented} and the Ensemble Kalman Filter (EnKF) \citep{evensen}.
Of all the computational difficulties associated with the standard Kalman filter, most of them 
can be traced back to the background covariance matrix. Indeed, this matrix is 
quadratic in the number of prediction points, which makes it hard to handle and store 
at larger problem sizes. One approach proposed to alleviate these difficulties is 
the ensemble Kalman filter (EnKF), first introduced by \citep{evensen} and subsequently 
developped in \citep{evensen_practical,evensen_book}. Ensemble Kalman 
filtering gets rid of the background covariance matrix by instead updating sequentially 
a sample (ensemble) of the prior background distribution and using the empirical sample 
covariance as an approximation of the true covariance. Thus, instead of storing a full covariance 
matrix, EnKF only has to store the (updated) ensemble members at each step.

For most applications, the EnKF provides an efficient, tractable approximation to the standard 
Kalman filter, but it has one significant drawback: for any application that requires interfering with the 
covariance, the difficulties of the standard Kalman filter reappear. 
An example of such an application is covariance regularization. Indeed, at tractable (small) 
ensemble sizes, the empirical covariance suffers from undersampling, which significantly 
degrades the performance of the filter due to spurious long-range correlations, 
variance underestimation, artifacts and other estimation errors. 
This means that, in practice, the empirical covariance has to be regularized before in can be used for filtering \citep{anderson_inflation,hamill_localization}, 
hence requiring access to the full matrix, bringing us back to the chores of the original filter.

In practice, these problems are usually solved by considering some sequential version of the filter \citep{houtekamer_sequential}, 
where observations are assimilated in batches, so that, at each iteration, only the parts of the covariance associated to the data points being currently 
assimilated need to be computed. It has long been known in the data assimilation community that, 
when used in conjunction with covariance regularization 
techniques, sequential processing breaks down the theoretical guarantees 
of the EnKF and can yield unwanted consequences. 
These include, among others, dependency of the results on observations ordering. 
Heuristical approaches for "optimal" observation 
ordering during assimilation have been proposed \citep{whitaker_ordering}, 
and while the problem is known, practical applications 
of the EnKF resort to sequential assimilation as a last resort, 
owing to the computational impossibility of assimilating 
all data synchronously, as would be required by the original filter formulation.

The first, and to our knowledge only, systematic study 
of order dependency in sequential ensemble Kalman filtering was undertaken by \citep{nerger_ordering}. 
This study demonstrates that the use of covariance localization \citep{hamill_localization} leads to observation ordering 
dependency when used in the square root ensemble Kalman filter of \citep{tippett_square_root}. The study concludes that 
when observational errors are similar in magnitude to model errors, the ordering effect is of moderate significance. 
While such an assumption can be met in some setups, it does not hold in general for ensemble-based climate reconstructions 
\citep{bhend_ensemble_climate} where model uncertainties can be large.

In this work, we introduce a new implementation of the ensemble Kalman filter that leverages the most recent 
progress in distributed computing. Through the use of distributed arrays \citep{dask,dask_review}, the full model error covariance 
matrix can be built in (distributed) memory, which allows our implementation to assimilate all observations in a single 
batch (all-at-once), thus getting rid of ordering dependencies. 
This allows to use localization in EnKF while 
still retaining the theoretical properties of the original filter. 
The computational scalability of this new implementation 
allows it to conduct all-at-once, localized, EnKF assimilation on large-scale problems such as global climate 
reconstruction \citep{bhend_ensemble_climate,twentieth_century_reanalysis,improved_twentieth_century,franke}. We believe that by leveraging the most recent developments 
in distributed computing, this new approach opens the door to a principled use of covariance localization in large-scale 
applications of ensemble (square-root) Kalman filtering.

To demonstrate our new implementation, we compare the performance of sequential assimilation and all-at-once assimilation 
on a synthetic problem and on a real-world paleoclimatic reconstruction application 
(20th century reanalysis). The climate reconstructions provided by the different assimilation 
techniques are compared using scoring rules \citep{gneiting_scoring}, which, 
though not broadly used in climatology, allow for a finer assessment 
of reconstruction quality by considering the full probabilistic nature of the assimilation compared 
to more traditional metrics that only treat the reconstruction as a pointwise prediction.

This paper is structured as follows: in \Cref{sec:kalman_filtering} we review the basics of the ensemble Kalman filter 
and its variants. In \Cref{sec:distributed} we introduce the distributed ensemble Kalman filter (dEnKF) and demonstrate 
its capabilities. Finally, in \Cref{sec:experiments}, we compare the performances of all-at-once assimilation with 
sequential assimilation in synthetic experiments as well as on real-world large-scale climate reconstruction problems.

Our study shows that all-at-once assimilation performs significantly better on all considered problems, while 
still being computationally tractable. This opens new venues for the study of the effects of covariance 
localization in climate reconstruction \citep{valler} and provides the first large-scale extension of the investigations 
on sequential vs synchronous assimilation performed in \citep{nerger_ordering} that were limited to small-scale 
setups, due to lack of computational resources.

\section{The Kalman Filter and its Variants}\label{sec:kalman_filtering}
Ensemble Kalman filtering (EnKF) is a data assimilation technique that aims at providing an 
estimation of the state of a dynamical system by combining observed data with a prior model of the unknown system state. 
Before going into the details of the EnKF, we review the classic (discrete) Kalman filter, 
adopting a Bayesian point of view for our exposition. The interested reader is referred to 
\citep{snyder_intro_kalman} and \citep{welch_intro_kalman} for more details. The exposition 
presented here is inspired from the one in \citep{katzfuss_enkf}.

In the following, let $\trueState_{\timeIndex} \in \stateSpace$ denote the unknown state vector of some physical system 
at time $\timeIndex$ and assume that the system is governed by the following dynamics
\begin{align}
    \trueState_{\timeIndex + 1} &= \dynamicFwd_{\timeIndex}\trueState_{\timeIndex} 
    + \modelNoise_{\timeIndex},
    ~ \modelNoise_{\timeIndex}\sim\mathcal{N}(0, \modelNoiseCov_{\timeIndex}),
\end{align}
where, for all $\timeIndex\in\mathbb{N}$, the dynamics of the system is determined by a linear 
operator $\dynamicFwd_{\timeIndex}:\stateSpace\rightarrow\stateSpace$ and 
$\modelNoise_{\timeIndex}$ is a random vector modelling 
our uncertainty in the dynamics of the system. In what follows, 
we always assume that $\modelNoise_{\timeIndex}$ is independent 
of $\modelNoise_{\timeIndex'}$ for $\timeIndex \neq \timeIndex'$.

At each time step $\timeIndex$, we are given observations that stem from 
a linear operator $\fwd_{\timeIndex}$ applied to the current unknown state of the system:
\begin{align}
    \dataVec_{\timeIndex} &= \fwd_{\timeIndex}\trueState_{\timeIndex} + \noise_{\timeIndex},
    ~
    \noise_{\timeIndex}\sim\mathcal{N}(0, \noiseCov_{\timeIndex})
    \label{eq:data_model},
\end{align}
where $\fwd_{\timeIndex}:\stateSpace\rightarrow \mathbb{R}^{\nData_{\timeIndex}}$ is a linear measurement operator 
and $\noise_{\timeIndex}$ is a random observation noise vector. 
We assume that $\noise_{\timeIndex}$ is independent of $\noise_{\timeIndex'}$ 
for $\timeIndex \neq \timeIndex'$. 
Kalman filtering aims at sequentially estimating (filtering) the state of 
the system, based on the observations available up to the current time step and 
on some prior knowledge about the initial state at time $0$. 
In a Bayesian formulation, Kalman filtering assumes that the initial state 
$\state_0$ is a realization of random vector $\randomState_0$ with 
Gaussian prior distribution $\randomState_0\sim\mathcal{N}(\mean_0, \cov_0)$ and 
then approximates the unknown state at time $\timeIndex$ using the 
posterior (filtering) distribution of $\randomState_0$ conditionally 
on the dynamics and the observations up to $\timeIndex$. This 
conditional distribution is Gaussian with mean and covariance that are 
given by:
\begin{align}
    \condMean_{\timeIndex} &= 
    \forecastMean_{\timeIndex} + \kalmanGain_{\timeIndex}\left(
        \dataVec_{\timeIndex} - \fwd_{\timeIndex}\forecastMean_{\timeIndex}
    \right),\\
    \condCov_{\timeIndex} &= \forecastCov_{\timeIndex} 
    - 
    \kalmanGain_{\timeIndex}\fwd_{\timeIndex}\forecastCov_{\timeIndex},
\end{align}
where the matrix $\kalmanGain_{\timeIndex}$ is called the \textit{Kalman gain} and is given 
by 
\begin{align}
\kalmanGain_{\timeIndex} = \forecastCov_{\timeIndex}\fwd^T_{\timeIndex}
\left(\fwd_{\timeIndex}\forecastCov_{\timeIndex}\fwd^T_{\timeIndex} 
+ \noiseCov_{\timeIndex}\right)^{-1}\label{eq:kalman_gain},
\end{align}
assuming the matrix in parenthesis to be invertible, and 
the \textit{forecast mean} $\forecastMean_{\timeIndex}$ and 
\textit{forecast covariance} $\forecastCov_{\timeIndex}$ are given by:
\begin{align*}
    \forecastMean_{\timeIndex} &=
    \dynamicFwd_{\timeIndex}\condMean_{\timeIndex-1},\\
    \forecastCov_{\timeIndex} &= \dynamicFwd_{\timeIndex}\condCov_{\timeIndex-1}\dynamicFwd^T_{\timeIndex} + \modelNoiseCov_{\timeIndex}.
\end{align*}
 
The key quantity to compute at each iteration is the \textit{Kalman gain}, which is 
a 
$\nModel \times \nData_{\timeIndex}$ matrix. 
In practice, the use of the above Kalman filter algorithm requires computing the $\nModel\times\nModel$ conditional 
covariance matrix $\condCov_{\timeIndex}$ at each step, 
which can be prohibitive or even impossible when the state 
space is large, as in the case, for example, in climate science applications. The Ensemble Kalman Filter (EnKF) was 
introduced to overcome these difficulties.

\subsection{Ensemble Kalman Filter}
The ensemble Kalman filter (EnKF) can be seen as a Monte-Carlo approximation 
of the classical Kalman filter that aims at bypassing the difficulties created 
by large conditional covariance matrices. 
Its key idea is to approximate the conditional distribution at each step 
by a sample (ensemble) drawn from the distribution. Then, instead 
of having to update mean vectors and covariance matrices, one only has to update 
the ensemble members, thereby reducing the computational burden and memory footprint.

Considering the same setting as in the previous section, 
ensemble Kalman filtering begins with an \textit{ensemble}, 
$\ensMember^{(1)}_{0},\dotsc, \ensMember^{(\nEns)}_{0} \overset{\textrm{i.i.d.}}{\sim}
\mathcal{N}(\condMean_{0}, \condCov_{0})$
which is a sample 
from the prior state distribution. At each step $\timeIndex\in\mathbb{N}$, the EnKF updates the 
current ensemble members 
$\ensMember^{(1)}_{t-1},\dotsc, \ensMember^{(\nEns)}_{t-1}$ to 
a new ensemble $\ensMember^{(1)}_{t},\dotsc, \ensMember^{(\nEns)}_{t}$ 
so that the new ensemble constitutes an i.i.d. sample of the conditional 
distribution at time $\timeIndex$. 
The traditional EnKF produces an ensemble that aproximates the conditional 
distribution by updating 
the mean and deviations using \textit{perturbed observations}. Indeed, 
as noted by \citep{burgers_perturbed}, one needs to inject randomness 
at each update step so that the filter does not underestimate the variance. 
The complete EnKF update procedure is described in \Cref{alg:EnKF_perturbed}. 
We note that the Gaussianity assumption on the ensemble can be dropped 
and refer the reader to \citep{grooms2022comparison} for a review of 
some non-Gaussian extensions of the EnKF.

\begin{algorithm}
    \caption{EnKF update (perturbed observations)}\label{alg:EnKF_perturbed}
\begin{algorithmic}
\Require Ensemble $\ensMember^{(1)}_{\timeIndex-1},\dotsc, \ensMember^{(\nEns)}_{\timeIndex-1}$, 
observation operator $\fwd_{\timeIndex}$, observed data $\dataVec_{\timeIndex}$ 
and observation noise covariance matrix $\noiseCov_{\timeIndex}$.
\Ensure Updated ensemble $\ensMember^{(1)}_{\timeIndex},\dotsc, \ensMember^{(\nEns)}_{\timeIndex}$.\\
\item[]
\begin{enumerate}
    \item Sample evolution errors 
        $\modelNoise^{(i)}_{\timeIndex}\overset{\textrm{i.i.d}}{\sim}
        \mathcal{N}(0, \modelNoiseCov_{\timeIndex})$.
    \item Forecast members
        \begin{align*}
            \ensMemberIf_{\timeIndex} &= \fwd_{\timeIndex}\ensMemberI_{\timeIndex - 1}
            + \modelNoise^{(i)}_{\timeIndex}.
        \end{align*}
    \item Sample  \textit{perturbed observations} 
$\dataVec_{\timeIndex}^{(i)}\overset{\textrm{i.i.d.}}{\sim} \dataVec_{\timeIndex}
+\noise_{\timeIndex}$
\item Estimate \textit{ensemble covariance matrix} $\estCov_{\timeIndex}$ and 
    compute \textit{Kalman gain}:
    $$
\kalmanGainEns_{\timeIndex} = \estCov_{\timeIndex}\fwd^T_{\timeIndex}
\left(\fwd_{\timeIndex}\estCov_{\timeIndex}\fwd^T_{\timeIndex} 
+ \noiseCov_{\timeIndex}\right)^{-1}
    $$
\item Update the ensemble 
members:
\begin{align}
    \ensMemberI_{\timeIndex} &= 
    \ensMemberIf_{\timeIndex}
    +
    \kalmanGainEns_{\timeIndex}\left(\dataVec_{\timeIndex}^{(i)} - \fwd_{\timeIndex}\ensMemberIf_{\timeIndex}\right)\label{eq:ens_updt_members}.
\end{align}
\end{enumerate}
\end{algorithmic}
\end{algorithm}
Here, compared to the traditional Kalman filter, one does not have access to the forecast 
covariance matrix $\forecastCov_{\timeIndex}$, so that one has to replace 
it by some estimator $\estForecastCov_{\timeIndex}$ computed from the ensemble. 
There exists various techniques for estimation of the forecast covariance, such 
as localization \citep{houtekamer_sequential,hamill_localization}, shrinkage \citep{ledoit_wolf,ledoit_wolf_nonlinear}, Bayesian estimation \citep{efron_morris,haff}, some 
of which are further explained in \Cref{sec:localization}.

One sees that, at each step, the EnKF maintains an ensemble of $\nEns$ state vectors that approximates 
the conditional distribution, and that the update equations are analogous to the classical 
Kalman update equations, with the state forecast covariance replaced by 
some ensemble estimate thereof and the observations replaced by perturbed ones.
The ensemble update equations \Cref{eq:ens_updt_members} were chosen 
so that, in the limit of infinite ensemble size, the empirical ensemble covariance converges to the true covariance. 
Although theoretically justified, the use of perturbed observations 
at every step is a source of sampling errors that reduces filter performances at small ensemble sizes 
(see \citep{whitaker_no_perturb} for a detailed study). Therefore, deterministic approaches that avoid 
the use of perturbed observations have been developed. Most of these approaches fall under the category 
of \textit{square root filters} \citep{tippett_square_root}, which we present next.

\subsection{Ensemble Square Root Filters} 
Ensemble square root filters (EnSRF) are a class of ensemble Kalman filters 
that use deterministic updates to avoid the use of perturbed observations 
\citep{whitaker_no_perturb,tippett_square_root}. As for the EnKF, our exposition 
of EnSRF will mostly follow the one in \citep{katzfuss_enkf} and we refer 
interested readers to this work for more details and to \citep{houtemaker_review} for 
a review of the different types of ensemble Kalman filters.

Denoting by $\ensMean_{\timeIndex}$ the mean of the ensemble members at 
time $\timeIndex$ and by 
$\ensDevI_{\timeIndex}:=\ensMemberI_{\timeIndex} - \ensMean_{\timeIndex}$ 
the deviations of the ensemble members from the mean, 
the EnSRF replaces the update equations of the EnKF by an update 
of the mean and an update of the deviations:
\begin{align}
    \ensMean_{\timeIndex} &= 
    \ensMeanForecast_{\timeIndex}
    + 
    \kalmanGainEns
    \left(\dataVec_{\timeIndex} - \fwd\ensMeanForecast_{\timeIndex}\right)\label{eq:ens_updt_mean_sr},\\
    \ensDevI_{\timeIndex} &= 
    \ensDevIf_{\timeIndex}
    - 
    \kalmanGainSR_{\timeIndex}\fwd\ensDevIf_{\timeIndex}\label{eq:ens_updt_members_sr},
\end{align}
where the \textit{square root Kalman gain} $\kalmanGainSR_{\timeIndex}$ is a matrix defined 
by:
\begin{align}
\kalmanGainSR_{\timeIndex} = \estForecastCov_{\timeIndex}\fwd^T_{\timeIndex}
\left(
\sqrt{
    \fwd_{\timeIndex}\estForecastCov_{\timeIndex}\fwd^T_{\timeIndex} 
+ \noiseCov_{\timeIndex}
}
\right)^{-1}
\left(
\sqrt{
    \fwd_{\timeIndex}\estForecastCov_{\timeIndex}\fwd^T_{\timeIndex} 
+ \noiseCov_{\timeIndex}
}
+ \sqrt{\noiseCov_{\timeIndex}}
\right)^{-1},
\label{eq:kalman_gain_SR}
\end{align}
and the mean and deviations are updated deterministically: 
$\ensMean_{\timeIndex} = \fwd_{\timeIndex}\ensMean_{\timeIndex}$, $\ensDevIf_{\timeIndex} = \fwd_{\timeIndex}\ensDevI_{\timeIndex - 1}$.
In the above, we assume the matrices in parenthesis to be invertible, and the square roots 
are allowed to be any matrix square root of the original matrix. 
The EnSRF has been found to overperform the EnKF with perturbed observations for small 
ensemble sizes \cite{whitaker_no_perturb}.

\subsection{Sequential Filtering and the Problem of Localization}\label{sec:localization}
When using ensemble Kalman filters, great care has to be taken in the choice 
of the ensemble forecast covariance estimator $\estForecastCov_{\timeIndex}$ that is 
used for the definition of both the \textit{ensemble Kalman gain} in the EnKF \Cref{eq:ens_updt_members} 
and of the \textit{square root Kalman gain} in the EnSRF \Cref{eq:kalman_gain_SR}. 
Indeed, in most applications, the ensemble size is drastically smaller than 
the state vector, so that estimators based on the empirical covariance of the ensemble 
will suffer from errors due to undersampling. Furthermore, the empirical covariance 
may fail to be positive-definite and thus fail to be invertible. The issues call 
for sophisticated estimation procedures for the forecast covariance matrix.

For spatial data assimilation problems, 
the undersampling errors are typically known to manifest themselves 
as so-called spurious correlations between 
distant locations. \textit{Localization} is a procedure to mitigate those 
and produce a better estimate of the ensemble covariance than the empirical one 
\citep{houtekamer_sequential,hamill_localization}. The idea behind localization 
is to taper the long-range correlations in the empirical covariance by using 
a distance-dependent smoothing. Formally, \textit{localization} estimates the 
ensemble forecast covariance matrix by
\begin{align*}
    \estForecastCov_{\timeIndex} &= \mathrm{Cov}\left(\left(
    \ensMemberIf_{\timeIndex}\right)_{i=1,\dotsc, \nEns}\right)\circ \locMatrix,
\end{align*}
where the first term denotes the empirical covariance of the ensemble forecast, 
$\locMatrix$ is a prescribed symmetric positive-definite \textit{localization matrix} and $\circ$ denotes elementwise product. 
The \textit{localization matrix} $\locMatrix$ is usually chosen according to the 
characteristics of the problem at hand. 
In applications where the state vector represents the values of some spatially varying field 
$f:\mathbb{R}^{\spaceDim}\times \mathbb{R}\rightarrow \mathbb{R}$, 
$(\state_{\timeIndex})_{j} = f(x_j, t)$, localization matrices are typically defined 
by applying a symmetric positive definite function $\locFun$ to the spatial locations 
so that $\locMatrix_{jk}=\locFun(x_j, x_k)$. We note that there also exists 
non-distance-based localization approaches, such as the ones developed by \citep{furrer}.

\subsubsection{Sequential Filtering}\label{sec:seq_filtering}
Even though the EnKF algorithm alleviates some of the computational burdens of the traditional 
Kalman filter by bypassing the update of the forecast covariance matrix, the implementation 
of the update equations \Cref{eq:ens_updt_mean_sr,eq:ens_updt_members_sr} is still 
fraught with potential memory overflows resulting from the computation of the 
square root Kalman gain matrix $\kalmanGainSR_{\timeIndex}$. Indeed, for 
a state space of dimension $\nModel$, the computation of the Kalman 
gain still requires building the  $\nModel\times\nModel$ estimated 
covariance matrix $\estCov_{\timeIndex}$, 
so that one quickly runs into bottlenecks for state spaces of moderate dimensions. 
One way to circumvent this is to assimilate the available observations sequentially, 
one at a time \citep{houtekamer_sequential}.

In this variant of the EnKF, at any given assimilation step $\timeIndex$, given a data vector $\dataVec_{\timeIndex}$ of size $\nData_{\timeIndex}$ 
and an observation operator $\fwd_{\timeIndex}$, one assimilates 
the individual data points $\dataVec_{\timeIndex, \dataIndex}$ sequentially by using the update 
equations \Cref{eq:ens_updt_mean_sr,eq:ens_updt_members_sr} for each observation and then 
using the updated ensemble as a starting point for the assimilation of the next observation. 
Note that such an assimilation scheme is only valid as long as the observations errors 
are uncorrelated. When assimilating only observation number $i$, the critical quantity 
to be computed is the product $\estCov_{t}(\fwd_{\timeIndex}^T)_{:\dataIndex}$, where $(\fwd^T)_{:\dataIndex}$ 
denotes the $\dataIndex$-th column of $\fwd_{\timeIndex}$. Being of the same dimension $\nModel$ 
as the state space, this product can easily be stored once computed 
and no further memory bottlenecks arise down the assimilation pipeline; nevertheless, 
the computation of the product still requires some care 
to handle the large matrix $\estCov_{\timeIndex}$. While there exists 
approaches that deal with more general observations, such as linear forms \citep{travelletti2021}, 
we here focus on the case where only pointwise observations of the state are available, 
i.e. $(\fwd_{\timeIndex})_{\dataIndex:}$ is $1$ at index $\stateInd_{\dataIndex}$ and $0$ elsewhere. 
In such settings, the product then involves only a single column of the estimated covariance 
matrix, drastically simplifying the computations. \Cref{alg:EnSRF_sequential} gives 
a detailed account of the assimilation procedure, assuming that at step $\timeIndex$ 
the available data is made of observations of the state vector at indices 
$\stateInd_1, ..., \stateInd_{\nData_{\timeIndex}}$, that is, the forward 
operator can be written as 
$\left(\fwd_{\timeIndex}\right)_{\dataIndex k} = \delta_{k \stateInd_{\dataIndex}}$ 
(note that for completeness we should add a time index to the $\stateInd_{\dataIndex}$'s, 
but we drop it for the sake of readability).

\begin{algorithm}
    \caption{EnSRF update (sequential version)}\label{alg:EnSRF_sequential}
\begin{algorithmic}
\Require Ensemble $\ensMember^{(1)}_{\timeIndex-1},\dotsc, \ensMember^{(\nEns)}_{\timeIndex-1}$, 
observation operator $\fwd_{\timeIndex}$, observed data $\dataVec_{\timeIndex}$ 
and observation noise covariance matrix $\noiseCov_{\timeIndex}$.
\Ensure Updated ensemble $\ensMember^{(1)}_{\timeIndex},\dotsc, \ensMember^{(\nEns)}_{\timeIndex}$.\\
\item[]
\begin{enumerate}
    \item Forecast members (deterministically)
        \begin{align*}
            \ensMemberIf_{\timeIndex} &= \dynamicFwd_{\timeIndex}\ensMemberI_{\timeIndex - 1}.
        \end{align*}
\item \textbf{for} each observation $\dataVec_{\timeIndex, \dataIndex}$ 
    and corresponding observation index $\stateInd_{\dataIndex}$ \textbf{do}:
    \begin{enumerate}[label*=\arabic*.]
        \item Estimate corresponding column $(\sampleCov{\timeIndex})_{:\dataIndex}$
            of the sample covariance.
        \item Apply localization 
            $(\estCov_{\timeIndex})_{:\dataIndex} = \locMatrix \circ 
            (\sampleCov{\timeIndex})_{:\dataIndex}$.
        \item Compute the single observation Kalman gains 
\begin{align*}
    \kalmanGain_{\timeIndex} &= (\estCov_{\timeIndex})_{\dataIndex:}
    \left((\estCov_{\timeIndex})_{\dataIndex\dataIndex} 
+ \noiseCov_{\timeIndex}\right)^{-1}\\
    \kalmanGainSR_{\timeIndex} &= (\estForecastCov_{\timeIndex})_{\dataIndex:}
\left(
\sqrt{
    (\estForecastCov_{\timeIndex})_{\dataIndex\dataIndex} 
    + (\noiseCov_{\timeIndex})_{\dataIndex\dataIndex}
}
\right)^{-1}
\left(
\sqrt{
    (\estForecastCov_{\timeIndex})_{\dataIndex\dataIndex} 
    + (\noiseCov_{\timeIndex})_{\dataIndex\dataIndex}
}
+ \sqrt{(\noiseCov_{\timeIndex})_{\dataIndex\dataIndex}}
\right)^{-1},
\end{align*}
            and update the ensemble members to get $\ensMemberI_{\timeIndex}$.
\item Use update as background for next assimilation:
\begin{align}
    \ensMemberIf_{\timeIndex} \gets \ensMemberI_{\timeIndex}.
\end{align}
\end{enumerate}
\end{enumerate}
\end{algorithmic}
\end{algorithm}

In the sequential EnSRF \Cref{alg:EnSRF_sequential}, localization is applied 
at every assimilation step. It turns out that this interplay of localization 
and sequential assimilation makes the update equations inconsistent, 
as was already noted by \citep{whitaker_no_perturb}. Indeed, in order to 
correctly implement sequential, localized ensemble square root Kalman filtering, 
one would need to first localize the whole sample covariance matrix 
and update the full matrix at every step, resulting in no computational savings 
compared to non-sequential assimilation.

While it is well known in the data assimilation community that 
the interplay of localization and 
sequential assimilation produces inconsistent results, little effort has been 
devoted to studying the magnitude of these inconsistencies and to fixing them. 
To the best of our knowledge, the only systematic study of this phenomenon 
has been undertaken by \citep{nerger_ordering}, which identified that, 
in such cases, assimilation results depend on observation ordering 
and that inconsistencies can become significant when 
observational errors are similar in magnitude to model errors. Our goal
in the rest of this work is to provide a distributed implementation of the 
EnSRF (\Cref{sec:distributed}) that allows for the assimilation of large datasets 
in a non-sequential fashion. 
By leveraging distributed computing, our \textit{all-at-once} (aao) assimilation 
algorithm is able to bypass the need for sequential processing and thus 
uses a consistent update scheme. Then, in \Cref{sec:experiments} we provide 
detailed comparisons of the performance of our all-at-once assimilation 
algorithm compared to sequential assimilation on both toy examples and real 
test cases.

\section{Distributed Non-Sequential Ensemble Kalman Filtering}\label{sec:distributed}
In order to overcome the memory bottlenecks that plague the ensemble Kalman filter, 
we turn to distributed computing techniques, with the goal of storing 
the full estimated state covariance matrix (which represents the main roadblock) 
in distributed memory. For our implementation, we use the \texttt{DASK} \citep{dask_review,dask} 
Python library, which is a framework that allows for seamless handling of 
distributed arrays and provides a chunked implementation of most traditional array 
algorithms.

In \texttt{DASK}, large arrays are chunked and stored split across several nodes in 
a computing cluster, while still providing a programming interface that allows for 
writing algorithms as if one were operating with a regular array. In a Kalman filtering 
context, this means that one is no longer limited by the storage of the state covariance 
matrix, as long as it can fit on the available computing cluster. 
We note that apart from storing the covariance matrix, there are two other operations 
in the EnSRF algorithm that are problematic when this matrix is large: 
computing its inverse and square root. Thankfully, inverses and square roots 
can be (approximately) computed in distributed memory by 
using the randomized SVD algorithm from \citep{halko2011}, which is 
implemented in \texttt{DASK} as \texttt{dask.array.linalg.compressed\_svd}.

We stress that the idea of using randomized SVD in ensemble Kalman filter has already been developed 
elsewhere, most notably in \citep{bopardikar,farchi_bocquet}. While our work shares this central idea with the ones just 
cited, our originality is threefold: i) we provide a \textbf{distributed} implementation in a user-friendly 
Python package, ii) we perform a 
\textbf{detailed study of the improvements} of all-at-once assimilation over the sequential one 
and iii) we demonstrate our algorithm on \textbf{large-scale} problems. On the other hand, 
the previously cited works do not provide a distributed implementation and do not compare themselves to sequential assimilation.

We now turn to the practical implementation of the EnSSRF algorithm in a distributed 
computing framework. 
In the following, let $\singVal_i$ and $\singVecL_i, i=1,\dotsc, \svdRank$ be the $\svdRank$-largest (approximate) 
singular values and corresponding left singular vectors of the estimated background covariance $\estCov$ delivered by the 
rank $\svdRank$ randomized truncated SVD algorithm of \citep{halko2011}. Then, one can compute approximate inverses and square roots 
of $\estCov$ via:
\begin{align*}
    \estCov^{-1} 
    &\approx 
\left[\singVecL_1, \dotsc, \singVecL_{\svdRank}
  \right]
\begin{pmatrix}
    1/\singVal_1 & & \\
    & \ddots & \\
    & & 1/\singVal_{\svdRank}
  \end{pmatrix}
  \left[\singVecL_1, \dotsc, \singVecL_{\svdRank}
  \right]^T\\
  \sqrt{\estCov} 
    &\approx 
  \left[\singVecL_1, \dotsc, \singVecL_{\svdRank}
  \right]
\begin{pmatrix}
    \sqrt{\singVal_1} & & \\
    & \ddots & \\
    & & \sqrt{\singVal_{\svdRank}}
  \end{pmatrix}
  \left[\singVecL_1, \dotsc, \singVecL_{\svdRank}
  \right]^T.
\end{align*}
Thanks to these approximations, one only has to store $\svdRank$ singular 
values and $\nModel$-dimensional singular vectors, providing substantial computational 
savings compared to the $\nModel^2$ needed for the full covariance matrix. 
We note furthermore that since the \texttt{DASK} library is based on lazy evaluation, 
the approximate inverse and square roots are never fully computed, only the 
relevant parts are built at runtime. Based on the above, let us define the 
procedures $\Call{ApproximateInverse}{(\singVal_i, \singVecL_i)_{i=1,\dotsc, \svdRank}}$ 
and $\Call{ApproximateSqrt}{(\singVal_i, \singVecL_i)_{i=1,\dotsc, \svdRank}}$, 
which, given the first $\svdRank$ 
approximate singular values and left singular 
vector 
$\singVal_i$ and $\singVecL_i, i=1,\dotsc, \svdRank$ of a matrix 
compute the approximate inverse, respectively the approximate square root 
of the matrix. On top of these, let us also define the procedure\\
$\Call{KalmanUpdate}{(\ensMember^{(i)}_{\timeIndex})_{i=1,\dotsc,\nEns},
\fwd_{\timeIndex}, \dataVec_{\timeIndex},
\estCov_{\timeIndex}^{-1}, \sqrt{\estCov_{\timeIndex}}}$, which performs the (ensemble square root) Kalman 
update \cref{eq:ens_updt_mean_sr,eq:ens_updt_members_sr}
for a given ensemble and observations, given precomputed inverses and square 
roots of the estimated state covariance matrix. Using these, 
we can now formulate an algorithm for a distributed implementation 
of a single step of the Ensemble Square Root Kalman Filter:
\begin{algorithm}
\caption{Distributed EnSRF update}\label{alg:EnSRF_distributed}
\begin{algorithmic}
\Require Ensemble $\ensMember^{(1)}_{\timeIndex-1},\dotsc, \ensMember^{(\nEns)}_{\timeIndex-1}$, 
observation operator $\fwd_{\timeIndex}$ and observed data $\dataVec_{\timeIndex}$\\
SVD cutoff rank $\svdRank$.
\Ensure Updated ensemble $\ensMember^{(1)}_{\timeIndex},\dotsc, \ensMember^{(\nEns)}_{\timeIndex}$.
\item[]
\State Build localized estimated covariance $\estCov_{\timeIndex}$ in distributed memory.
\item[]
\State $(\singVal_i, \singVecL_i)_{i=1,\dotsc, \svdRank} \gets \Call{DistributedSVD}{\estCov_{\timeIndex}, \textrm{rank}=\svdRank}$
\State $\estCov_{\timeIndex}^{-1} \gets \Call{ApproximateInverse}{(\singVal_i, \singVecL_i)_{i=1,\dotsc, \svdRank}}$
\State $\sqrt{\estCov_{\timeIndex}} \gets \Call{ApproximateSqrt}{(\singVal_i, \singVecL_i)_{i=1,\dotsc, \svdRank}}$
\item[]
\State $(\ensMember^{(i)}_{\timeIndex})_{i=1,\dotsc,\nEns} \gets 
\Call{KalmanUpdate}{(\ensMember^{(i)}_{\timeIndex})_{i=1,\dotsc,\nEns},
\fwd_{\timeIndex}, \dataVec_{\timeIndex},
\estCov_{\timeIndex}^{-1}, \sqrt{\estCov_{\timeIndex}}}$
\Comment{Update using \cref{eq:ens_updt_mean_sr,eq:ens_updt_members_sr}}
\end{algorithmic}
\end{algorithm}

When implemented in a lazy-evaluation framework such as \texttt{DASK}, the memory 
requirements of \Cref{alg:EnSRF_distributed} are twofold: first there should be enough 
memory to build the estimated state covariance matrix (localized) $\estCov_{\timeIndex}$ 
second there should be enough memory left to compute the approximate SVD of this 
matrix. Once this is done, memory can be cleared and only the firs $\svdRank$ singular 
values and left singular vectors need to be stored. 
In the following section, we demonstrate how our distributed implementation of 
the EnSRF algorithm allows for non-sequential (all-at-once) assimilation 
in large-scale problems and show how it outperforms sequential assimilation 
(which uses inconsistent update equations, see \Cref{sec:seq_filtering}) on both a synthetic test case and a 
paleoclimatic reconstruction problem.

\section{Experimental Comparison of Sequential and Non-Sequential Filtering}\label{sec:experiments}
Even if non-sequential ensemble Kalman filtering is theoretically guaranteed 
to overperform its sequential counterpart when the assumptions of the Kalman filter are 
met; practical applications of the EnKF almost always lie outside of the validity range of 
these assumptions. We thus here provide numerical comparisons between all-at-once (aao) 
and sequential (seq) assimilation schemes for the EnKF on practical assimilation problems.
In order to compare performances, 
we run the two schemes in a variety of different scenarios where ground-truth for the state to be estimated 
is available. We assess the reconstruction quality of the approaches using three different metrics:
\begin{itemize}
    \item Root mean-square error (RMSE) when predicting with the analysis mean,
    \item RMSE skill score \citep{re_score},
    \item Energy score (ES) multivariate scoring rule \citep{gneiting_energy_score}.
\end{itemize}
While the first two are commonly used metrics in the Kalman filter community, 
scoring rules have yet to see broad application in those fields. 
Compared to the other metrics, the energy score takes the full analysis ensemble into 
account and also allows for theoretically sound ranking of different models. 
These scoring rules are explained in more details in the coming sections.

\subsection{Synthetic Test Case}\label{sec:synthetic}
We first compare the all-at-once (aao) and sequential (seq) implementation of the EnKF on 
a synthetic test case. The task is to reconstruct a two-dimensional scalar field on the unit square, 
discretized on a $\nGrid \times \nGrid$ square grid. The ground truth is generated by sampling a Mat\'ern $3/2$ Gaussian process (GP) with 
unit variance and correlation 
length $\lambda= 0.1$. A starting ensemble (background) is also generated by sampling $\nEns$ realizations from the same process, putting ourselves 
in a "well-specified" setup. 
All-at-once and sequential implementations are then compared by assimilating $\nData$ data points at randomly chosen locations.

\begin{figure}[h!]
\centering
  \includegraphics[width=0.45\linewidth,height=0.4\linewidth]{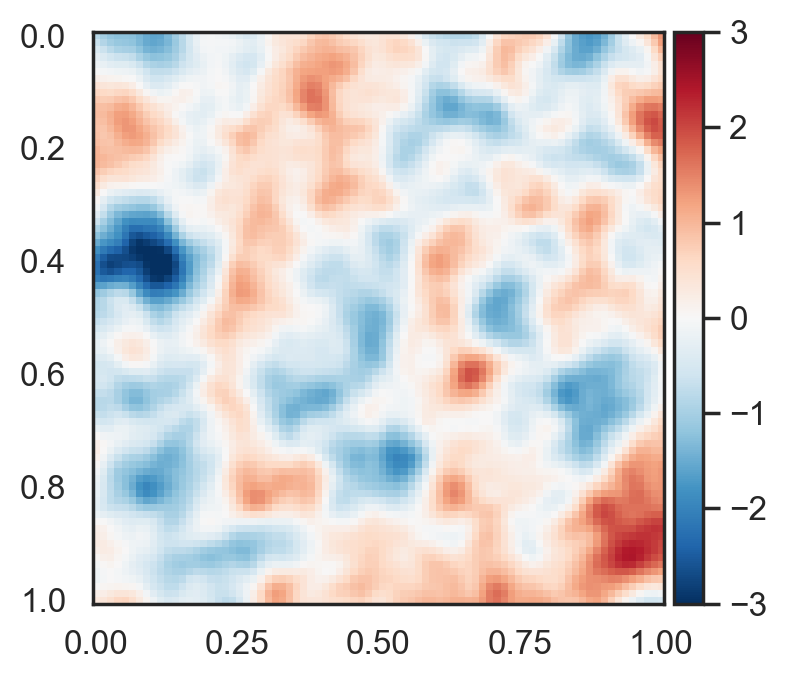}
  \caption{Example of ground truth sampled from the GP model.}\label{fig:ground_truth_synthetic}
\end{figure}

\begin{figure}[h!]
\begin{subfigure}{0.32\textwidth}
	\includegraphics[width=0.99\linewidth]{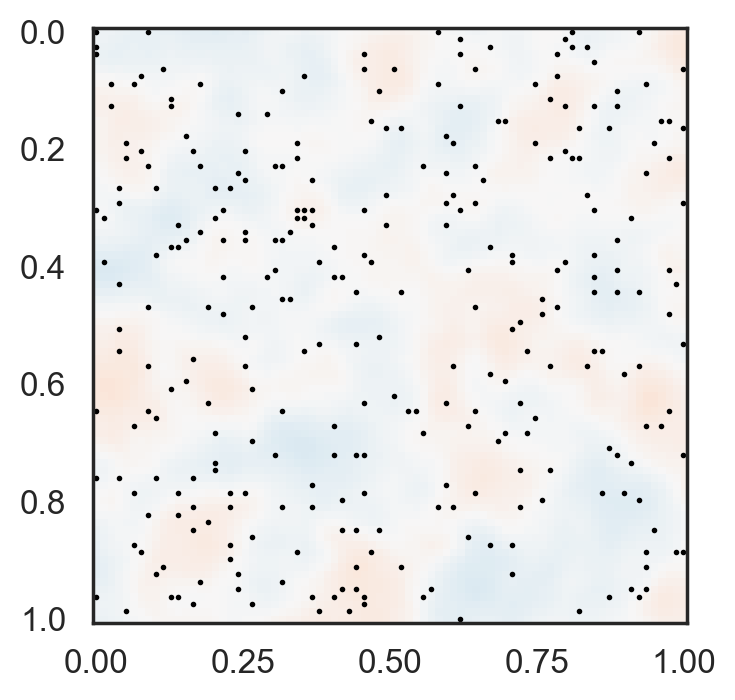}
	\caption{prior\\RMSE: 0.808}
\end{subfigure}
\begin{subfigure}{0.32\textwidth}
	\includegraphics[width=0.99\linewidth]{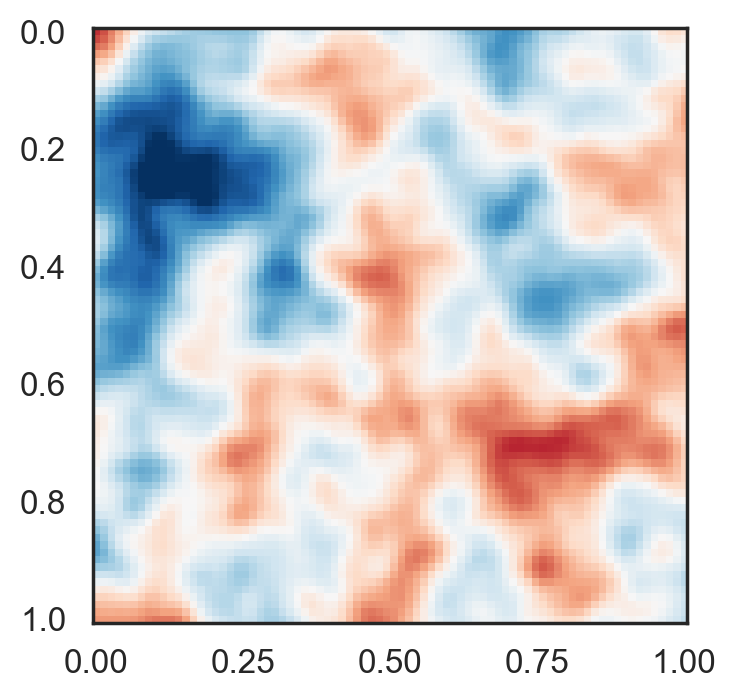}
	\caption{member 1\\RMSE: 1.283}
\end{subfigure}
\begin{subfigure}{0.32\textwidth}
	\includegraphics[height=0.965\linewidth]{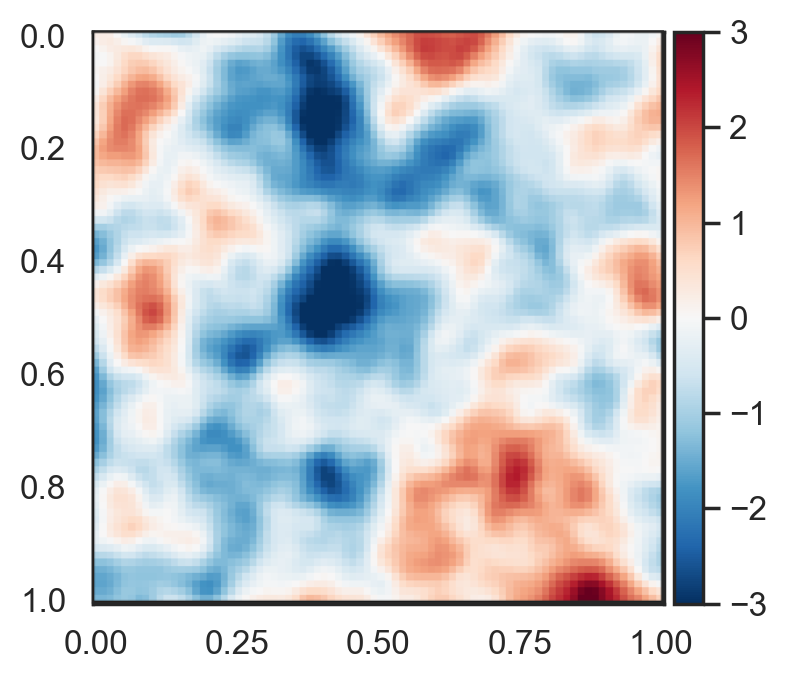}
	\caption{member 2\\RMSE: 1.315}
\end{subfigure}
\caption{Prior mean (a) and selected prior ensemble members (b) and (c). Observation locations depicted in black. RMSE for prediction 
of the ground truth in \cref{fig:ground_truth_synthetic} also depicted for each member.}
\label{fig:synthetic_prior}
\end{figure}

To compare the two assimilation schemes, a set of $20$ different ground truths is generated by sampling 
from the aforementioned GP model. For each such realization, a starting ensemble of size $30$ is also sampled from the same GP model, 
the number of ensemble members being chosen such as to reproduce situations encountered in practice. 
Then, a set of $\nData=1000$ randomly chosen data points is assimilated using each scheme and the assimilation results are compared using 
the aforementioned metrics. The ground truth are discretized on a $80\times 80$ square grid and the goal is to reconstruct 
the discretized ground truth vector. \Cref{fig:ground_truth_synthetic} and \cref{fig:synthetic_prior} depict the starting setup for one iteration 
of the experiments. The observational noise standard deviation is set to $\noiseStd = 0.01$, which corresponds to $1\%$ of the standard deviation of 
the GP model used to generate the ground truths and the ensembles. The effect of the observational noise level on the assimilation is further studied in 
follow-up experiments.

The covariance localization used in the experiment was composition with a symmetric positive definite function (see \cref{sec:localization}). The localization function used 
was a Mat\'ern $3/2$ kernel \citep{rasmussen_williams} with correlation length set to $0.2$ (twice the one of the GP used to generate the ground truths). We note that this choice 
is arbitrary but was made to strike a compromise between realism (one does not know the ''right`` localization function in practice) and reconstruction quality.

\begin{figure}[t]
    \captionsetup[subfigure]{justification=centering}
    \centering
    \begin{subfigure}[t]{0.3\textwidth}
        \makebox[0pt][r]{\makebox[30pt]{\raisebox{40pt}{\rotatebox[origin=c]{90}{\textbf{sequential}}}}}%
    \includegraphics[width=0.99\textwidth]{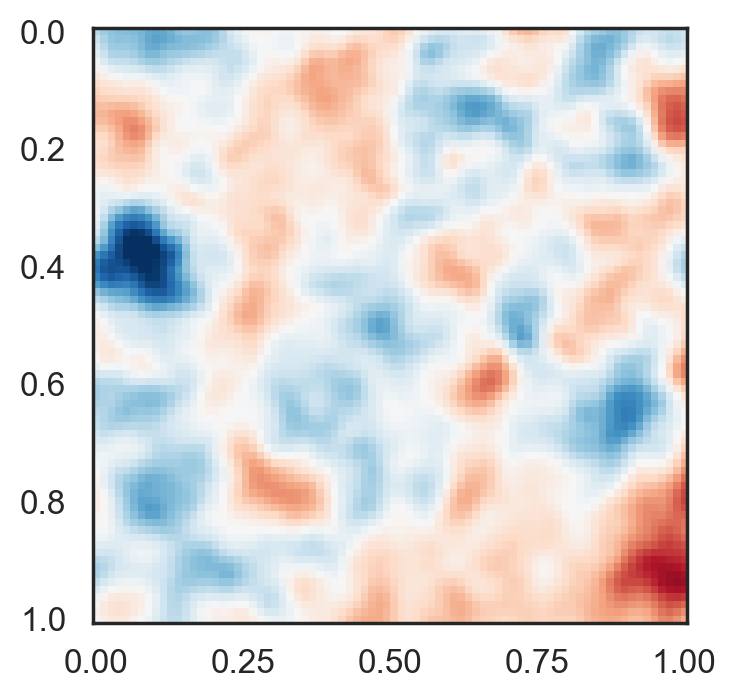}
    \caption{mean\\RMSE=0.361\label{fig:mean_updated_seq_synth}}
    \makebox[0pt][r]{\makebox[30pt]{\raisebox{40pt}{\rotatebox[origin=c]{90}{\textbf{all-at-once}}}}}%
	\includegraphics[width=0.99\textwidth]{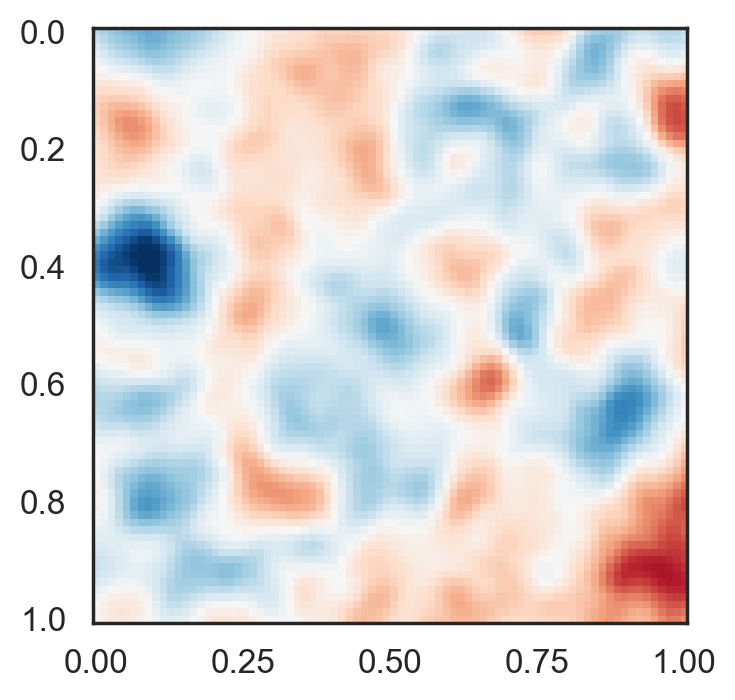}
    \caption{mean\\RMSE=0.322}\label{fig:mean_updated_aao_synth}
\end{subfigure}
\hspace{1em}
\begin{subfigure}[t]{0.3\textwidth}
	\includegraphics[width=0.99\textwidth]{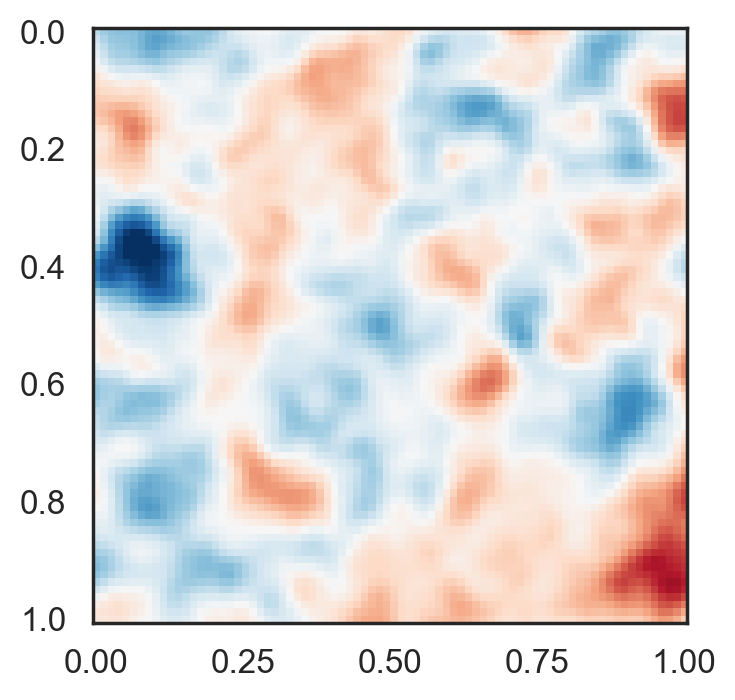}
    \caption{member 1\\RMSE=0.371}
	\includegraphics[width=0.99\textwidth]{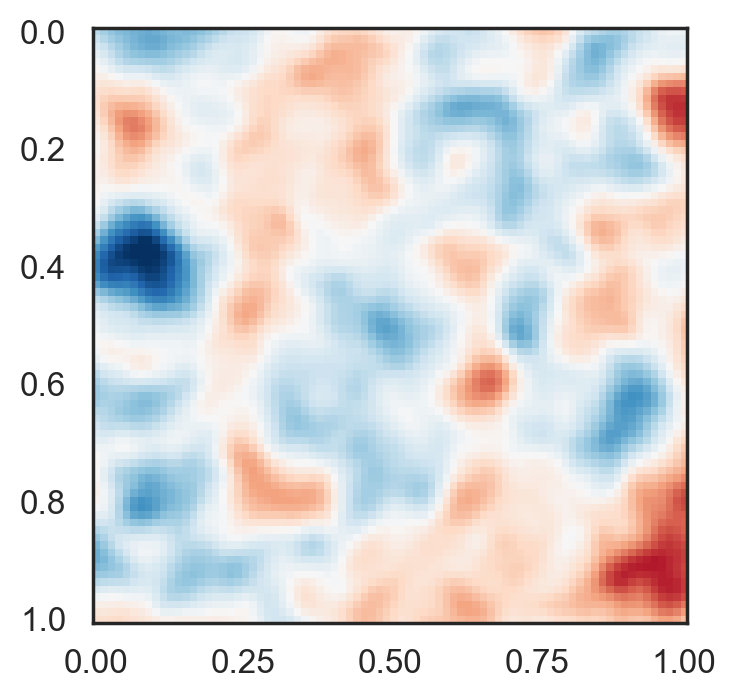}
    \caption{member 1\\RMSE=0.346}
\end{subfigure}
 \hspace{1em}
 \begin{subfigure}[t]{0.3\textwidth}
 	\includegraphics[height=0.957\textwidth]{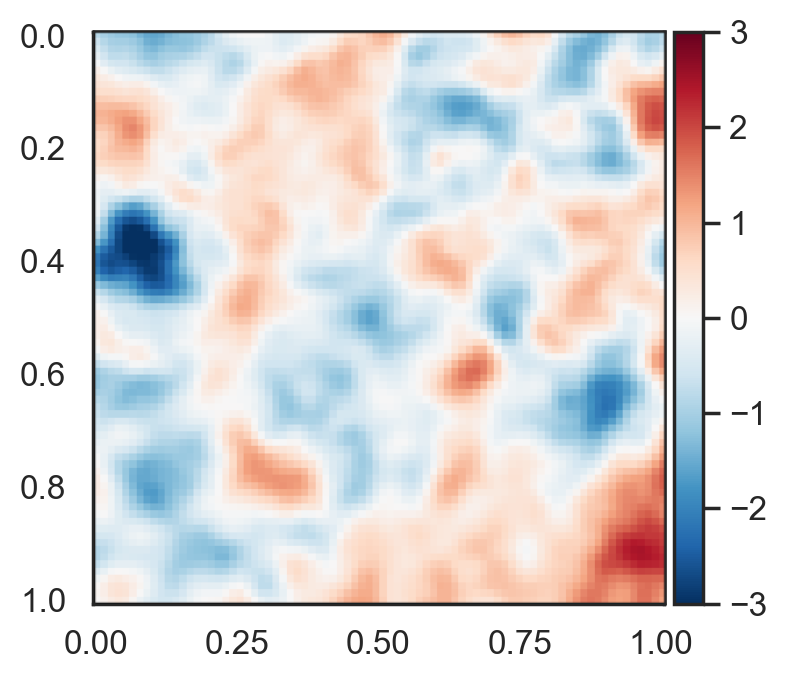}
    \caption{member 2\\RMSE=0.364}
    \vspace{-0.04cm}
 	\includegraphics[height=0.957\textwidth]{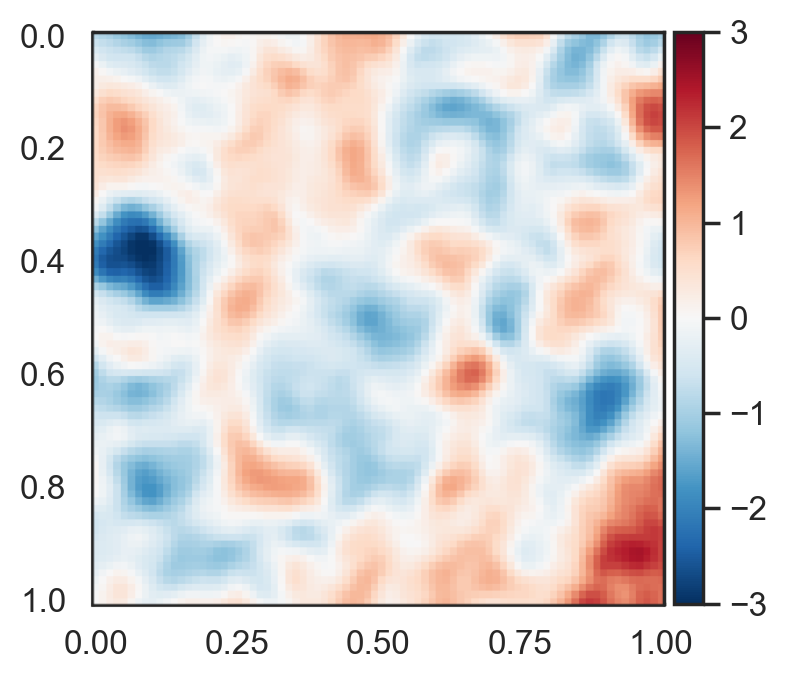}
    \caption{member 2\\RMSE=0.361}
\end{subfigure}
\caption{Updated means and selected updated ensemble members, 
after assimilation of the data generated from the ground truth in \Cref{fig:ground_truth_synthetic}.}
\label{fig:synthetic_updated}
\end{figure}

Results for assimilation of the data corresponding to the ground truth from 
\Cref{fig:ground_truth_synthetic} are depicted in \Cref{fig:synthetic_updated}. 
No obvious qualitative differences between the two assimilation schemes are to be found. 
Quantitative differences can be computed by considering, 
the root-mean-square error (RMSE) when reconstructing the ground truth using 
the data assimilation results. The RMSE when predicting the ground truth with the updated mean 
and two selected ensemble members is also depicted in \Cref{fig:synthetic_updated} for each 
assimilation scheme. One sees that all-at-once assimilation tends to perform better 
than the sequential approach.

\begin{figure}[h!]
\begin{subfigure}{0.45\textwidth}
	\includegraphics[width=0.99\linewidth]{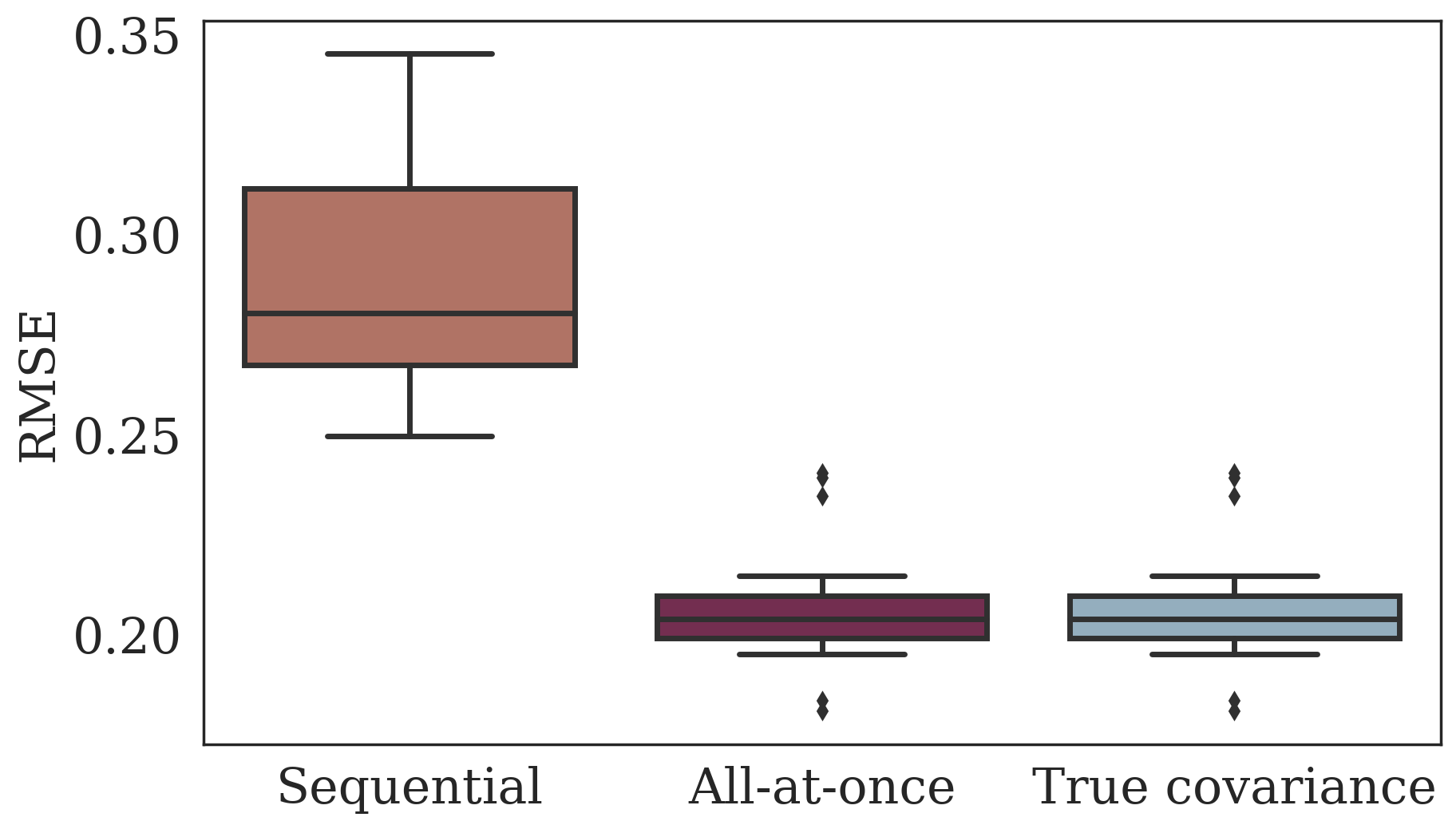}
	\caption{RMSE}
\end{subfigure}
\begin{subfigure}{0.45\textwidth}
	\includegraphics[width=0.99\linewidth]{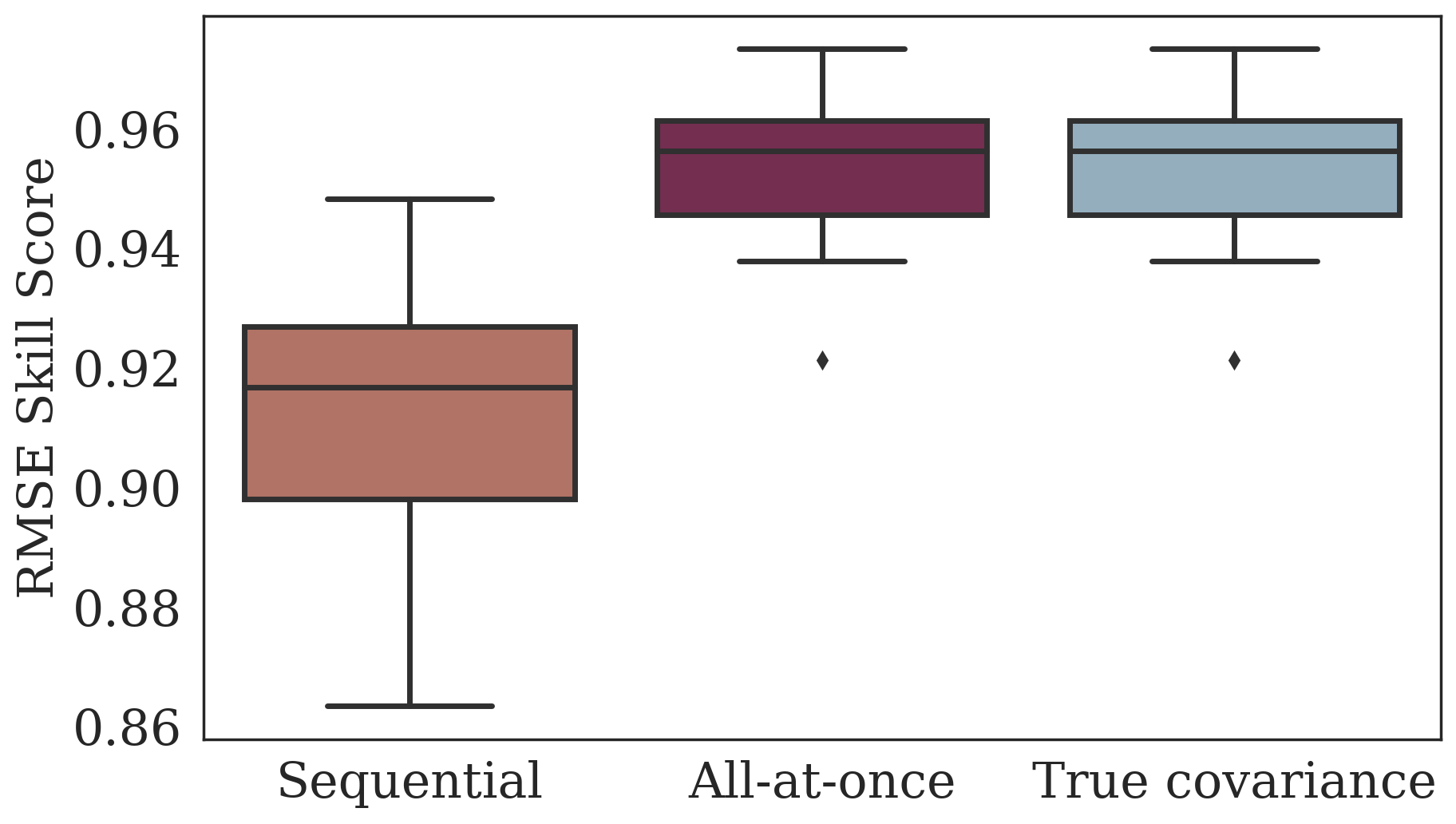}
	\caption{RMSE skill score}
\end{subfigure}
\caption{Comparison of the distributions (20 repetitions) of the RMSE and RMSE skill score (\textbf{synthetic test case}).}\label{fig:comp_rmse_re_synth}
\end{figure}

%
In order to get a quantitative assesment of the difference in accuracy between sequential 
and all-at-once assimilation, we repeat the assimilation experiments 20 times by sampling 
different ground truth from the GP model described before. Apart from the reconstruction RMSE, 
one can also compute the difference in the \textit{RMSE skill score}
for the different schemes. The \textit{RMSE skill score} \citep{re_score_murphy} 
is a widely used metric for ranking model performances in atmospheric science. For a given set 
of forecast vectors $\state^f_{\timeIndex}$, the RMSE skill score aims at assessing how well the 
forecast improves the reconstruction a given reference $\state^r_{\timeIndex}$ compared 
to some original background forecast $\state^b_{\timeIndex}$, 
over some time period $\timeIndex\in T$. The RMSE skill score is given by 
(time index ommited for brevity):
\begin{align*}
    \mathrm{RMSEss}\left(\state^f, \state^r, \state^b\right) 
    :&=
    1 - \frac{\sum_{\timeIndex\in T}||\state_{\timeIndex}^f - \state_{\timeIndex}^r||^2}{
    \sum_{\timeIndex\in T}||\state_{\timeIndex}^b - \state_{\timeIndex}^r||^2}
\end{align*}
Note that while the original definition of the RMSE skill score does not include a time 
dimension, it is customary in climate science to sum over the considered time steps. 
Compared to the RMSE, this metric weighs the errors by the best improvement that can be achieved, 
i.e. errors where the background is far from the reference are penalized less heavily. 
The RMSE skill 
score ranges from $1$ to $-\infty$, with a value of $1$ corresponding to perfect reconstruction. 
\Cref{fig:comp_rmse_re_synth} shows the distribution of the RMSE and RMSE skill score for sequential and 
all-at-once assimilation. Since these results depend on the choice of localization used, 
we also include the results for all-at-once assimilation using the true covariance matrix 
that was used to generate the ground truth. This gives an idea of the best possible reconstruction 
that can be achieved under the current conditions.

We see that all-at-once outperforms sequential assimilation for all metrics by an 
order of $2$ to $5\%$ (see \citep{wheatcroft_skill_score} on interpreting skill scores).

\begin{figure}[h!]
    \centering
	\includegraphics[width=0.6\linewidth]{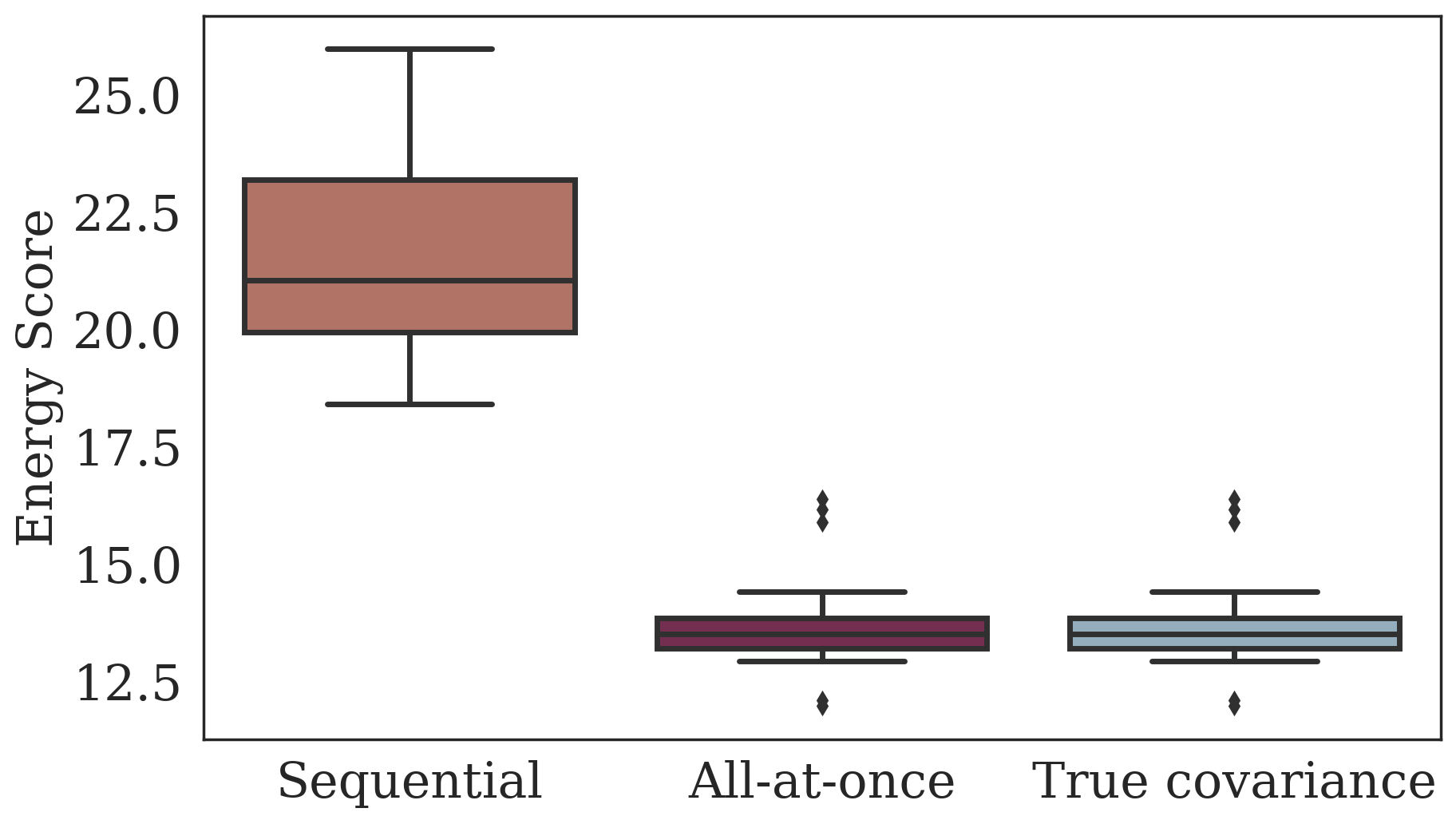}
    \caption{Comparison of the energy score for the different assimilation methods (\textbf{synthetic test case)}.}
\label{fig:comp_es_synth}
\end{figure}

While the RMSE skill score is widely used in climate science, it has been pointed out 
in the literature \citep{wheatcroft_skill_score} that skill scores can 
be misleading and biased compared to raw scores. Also, we would like to stress that, 
in the context of ensemble Kalman filters,
the RMSE skill score does not leverage the full content of the forecast since it only 
considers pointwise prediction (ensemble mean) when we have a full ensemble at hand. 
In order to incorporate the probabilistic information contained in the spread 
of the forecast ensemble in the ranking of the assimilation schemes, one should 
use \textit{proper scoring rules} \citep{gneiting_scoring}. 
We here only focus on the \textit{energy score} (ES) \citep{gneiting_energy_score}, 
which is a proper scoring rule for multivariate forecasts. 
For a given ensemble of forecast vectors $\state^f_{\timeIndex}, i=1,\dotsc, \nEns$ 
and a reference vector $\state^r_{\timeIndex}$ to be reconstructed, 
the energy score is given by \citep{jordan_scoring}:
\begin{align*}
    \mathrm{ES}\left(\ensMemberIf_{\timeIndex}, \state^r_{\timeIndex}\right)
    :&=
    \frac{1}{\nEns}\sum_{i=1}^{\nEns}
    ||\ensMemberIf_{\timeIndex} - \state^r_{\timeIndex}||
    - \frac{1}{2\nEns^2}\sum_{i=1}^{\nEns}\sum_{j=1}^{\nEns}
||\ensMemberIf_{\timeIndex} - \ensMemberJf_{\timeIndex}||
\end{align*}
Scoring rules are used for model comparison and ranking, lower scores corresponding 
to better models, according to the scoring at hand. 
While the energy score has some shortcomings (see \citep{bjerregard_multivariate_scoring} 
for a review and presentation of alternative scoring rules), 
such multivariate scoring rules have yet to find broad use in climate science 
and we firmly believe that the multivariate and probabilitsic nature of ES 
already offers an improvement over the univariate point-prediction scorings 
used in the community. \Cref{fig:comp_es_synth} shows that all-at-once also outperforms 
sequential assimilation in terms of energy score.

\begin{figure}[h!]
\begin{subfigure}{0.33\textwidth}
    \includegraphics[width=0.99\linewidth]{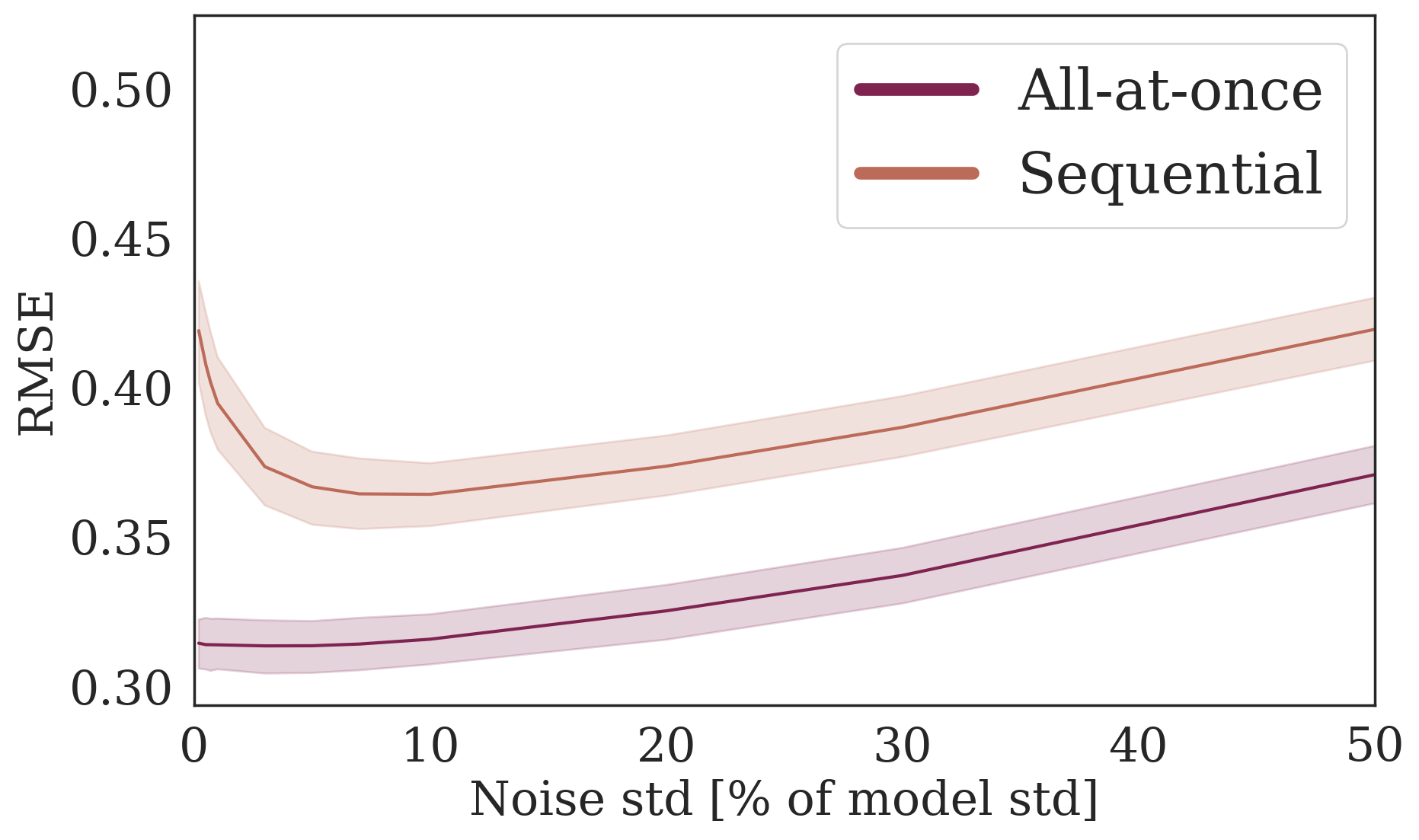}
	\caption{RMSE}
\end{subfigure}
\begin{subfigure}{0.33\textwidth}
    \includegraphics[width=0.99\linewidth]{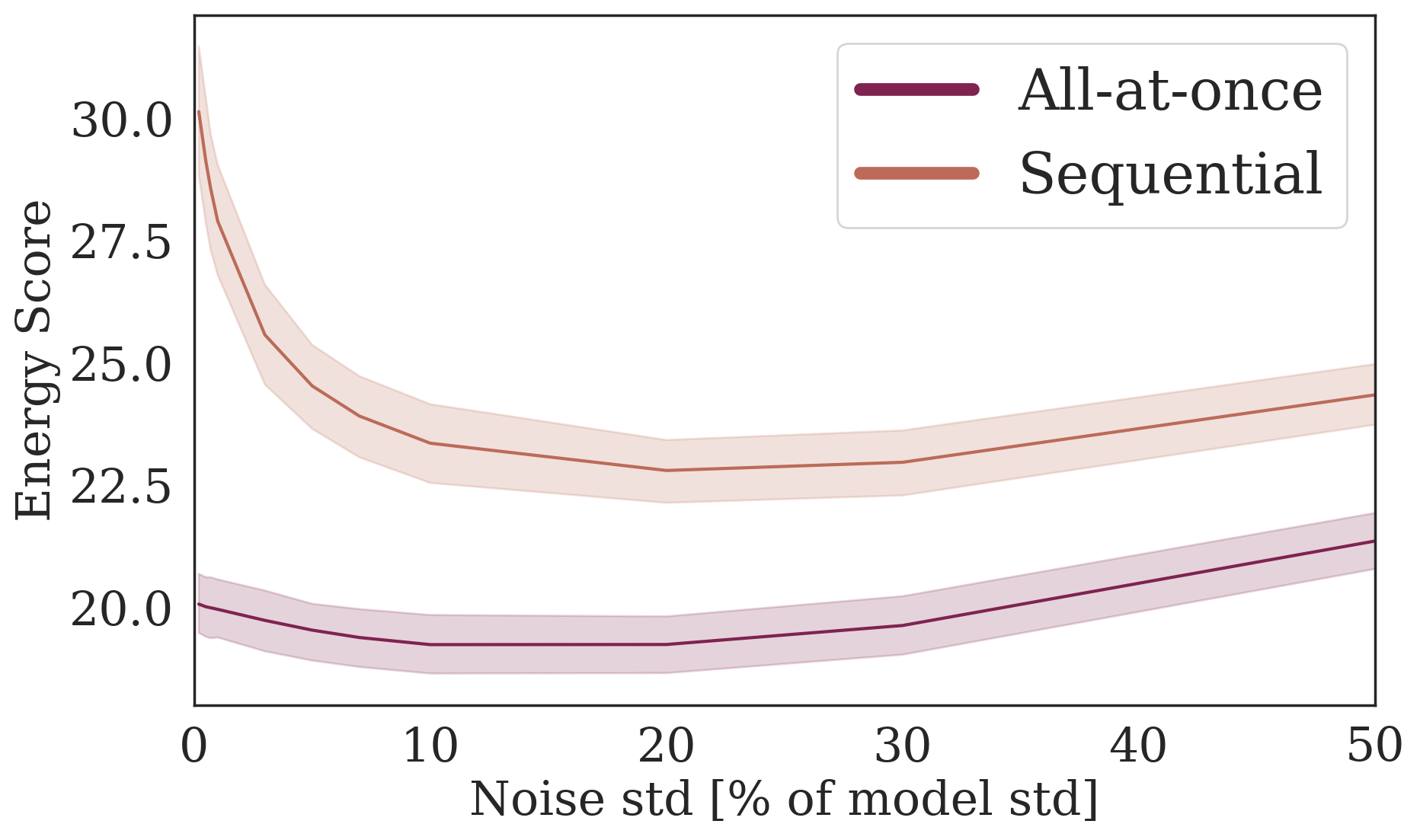}
	\caption{Energy score}
\end{subfigure}
\begin{subfigure}{0.33\textwidth}
    \includegraphics[width=0.99\linewidth]{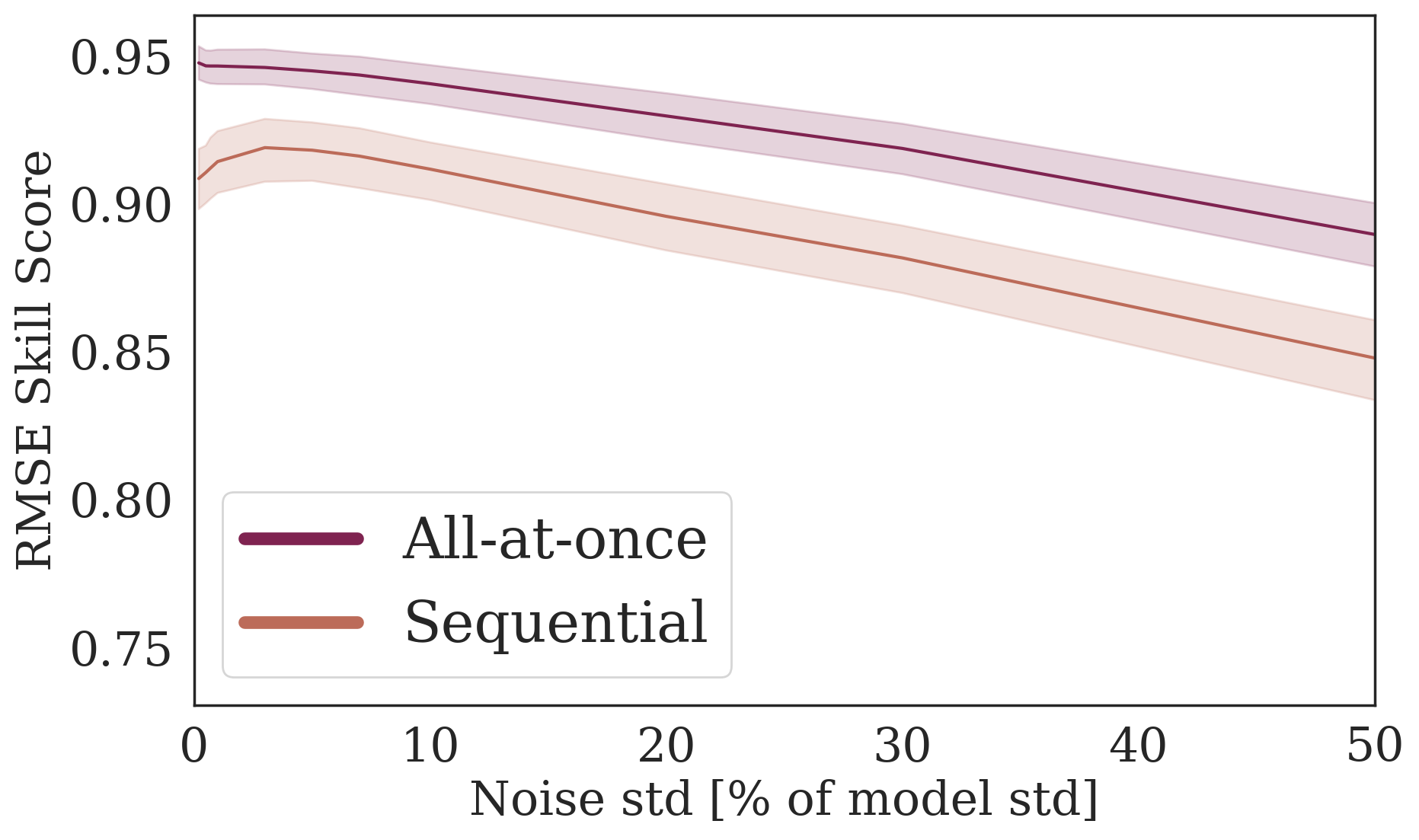}
	\caption{RMSE skill score}
\end{subfigure}
\caption{Evolution of the different accuracy metrics as a function of the observational noise standard deviation (\textbf{synthetic test case}).}
\label{fig:synthetic_scores_evolution}
\end{figure}

The amount by which all-at-once and sequential assimilation differ for the EnSRF 
depends on the magnitude of the observation errors compared to the spread of the 
starting ensemble. Indeed, as noted by \citep{nerger_ordering} 
\textit{``When the observation errors [are] of a similar magnitude as 
the initial errors of the state estimate, both
filter methods [show] a similar behavior. When the
observation errors [are] decreased, the EnSRF [shows]
a stronger tendency to diverge and larger minimum
RMS errors [...] than variant of the
EnSRF that assimilates all observations at once.''} 
\Cref{fig:synthetic_scores_evolution} demonstrates this effect 
by increasing the observational error standard deviation $\noiseStd$. 
Degradation of performance for the sequential EnSRF at small observational errors 
is clearly visible.

\begin{figure}[h!]
\begin{subfigure}{0.32\textwidth}
	\includegraphics[width=0.99\linewidth]{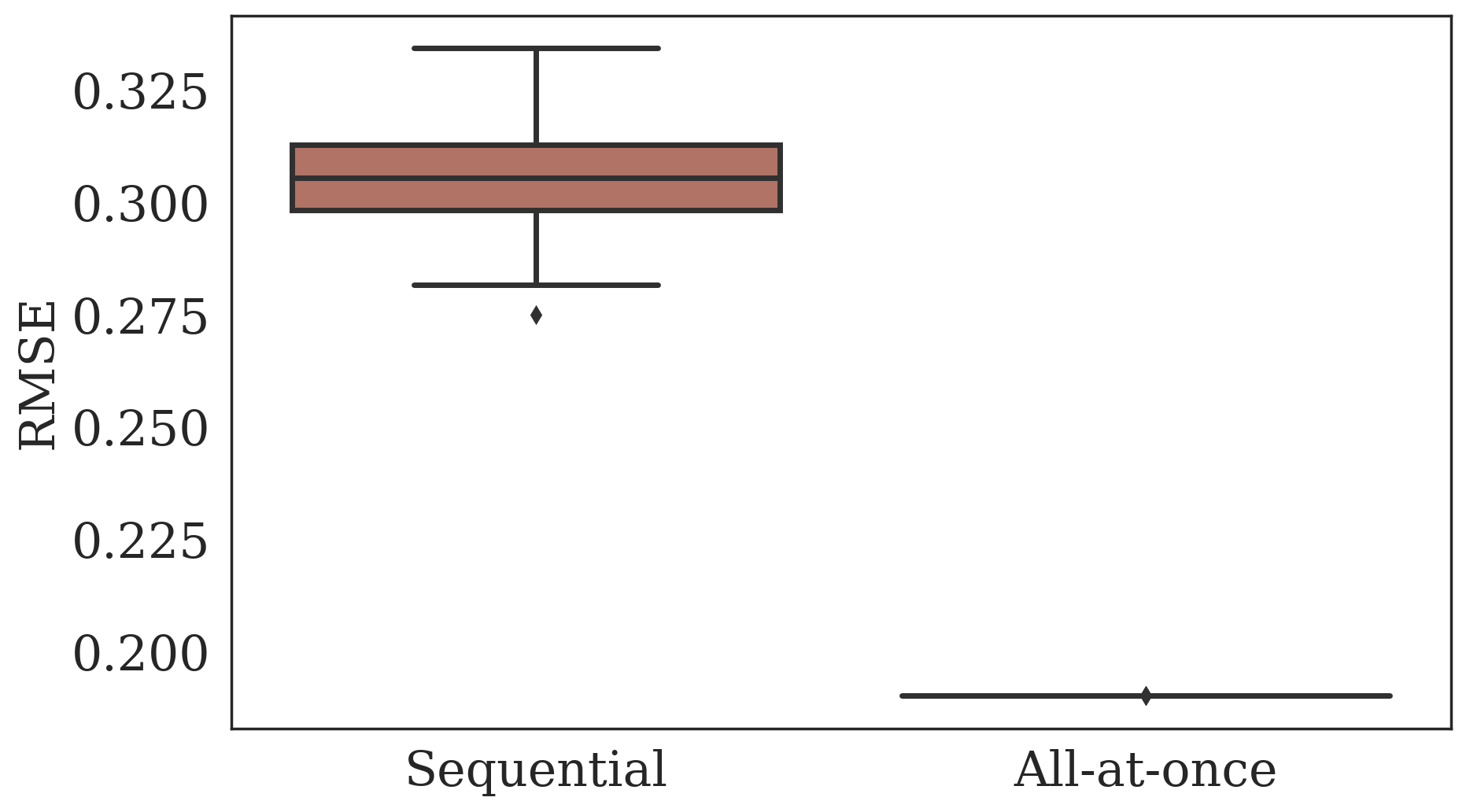}
	\caption{RMSE}
\end{subfigure}
\begin{subfigure}{0.32\textwidth}
	\includegraphics[width=0.99\linewidth]{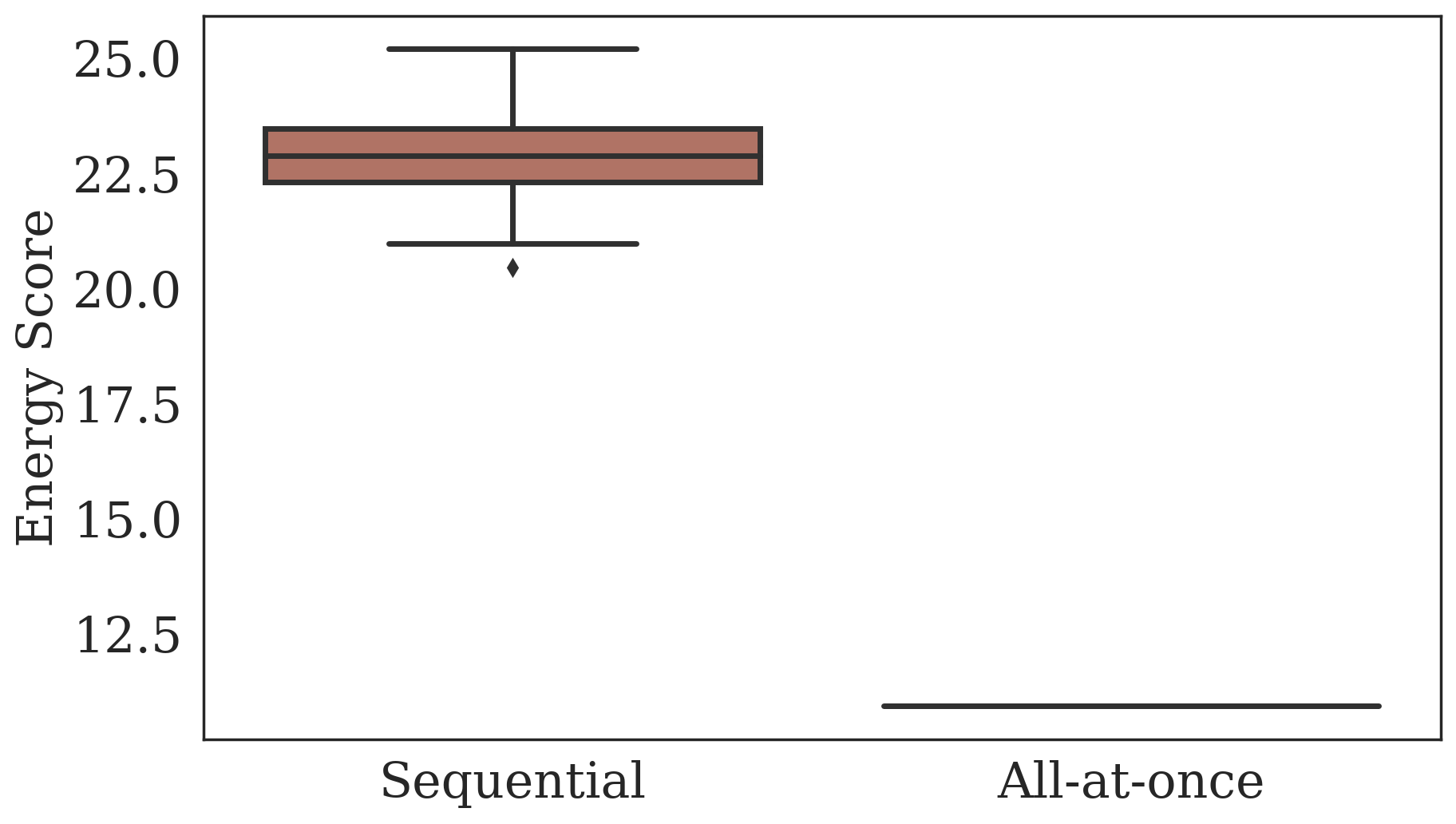}
	\caption{Energy score}
\end{subfigure}
\begin{subfigure}{0.32\textwidth}
	\includegraphics[width=0.99\linewidth]{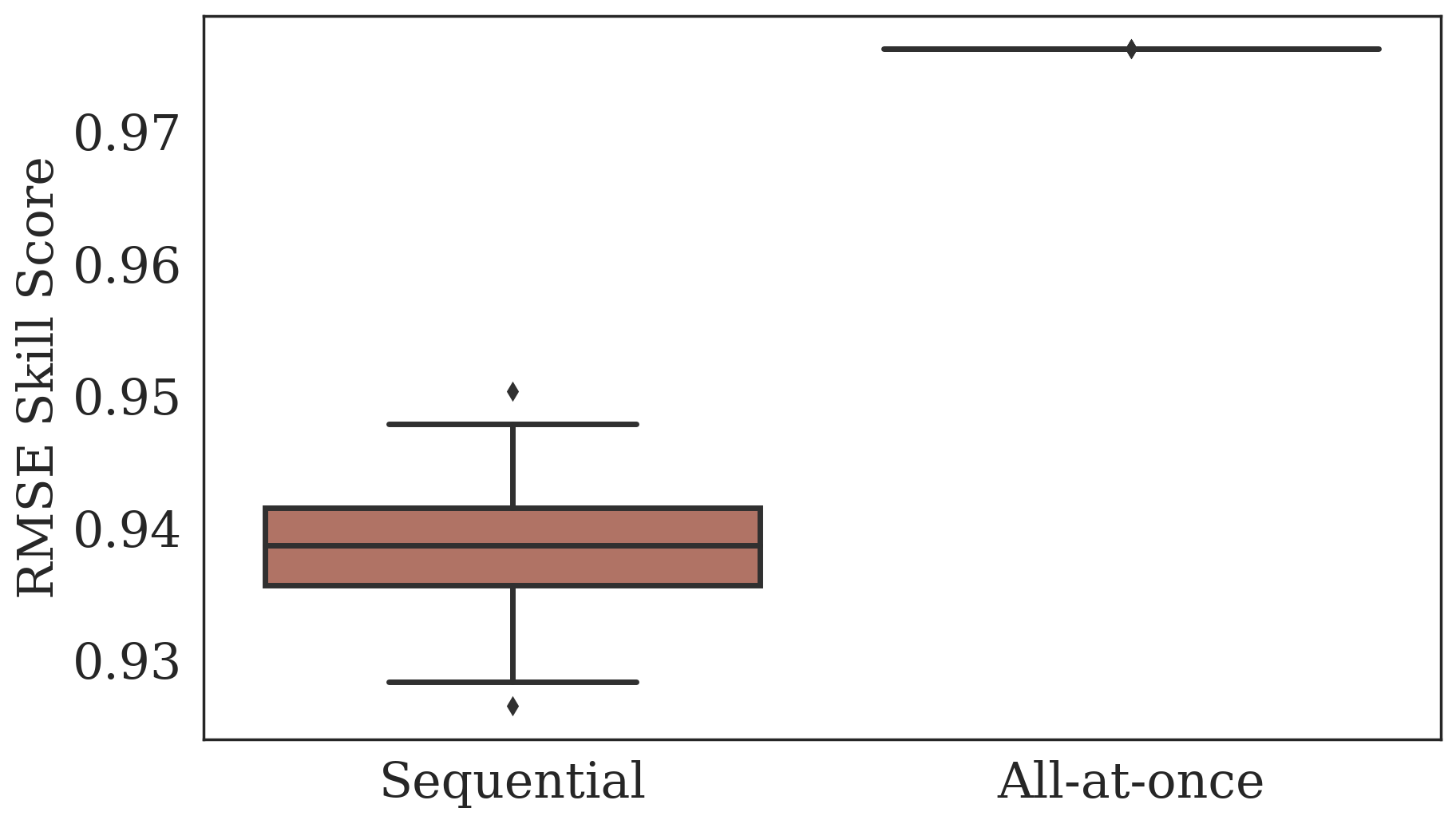}
	\caption{RMSE skill score}
\end{subfigure}
\caption{Comparison of the distributions of the accuracy metrics for different observation orderings using sequential assimilation (\textbf{synthetic test case}).}
\label{fig:synthetic_scores}
\end{figure}

On top of the above effect, incorrect updates in the localized, sequential version 
of the EnSRF introduce dependencies on the order of the observations, 
as was also noted by \citep{nerger_ordering}. While this effect is well known 
in the data assimilation community, 
it is usually assumed to be of little practical relevance. Apart from the small-scale 
experiments (40 observations) performed by \citep{nerger_ordering}, evidence for this 
assumption is lacking. To study the effect of ordering in larger assimilation setups, 
we use the same setting as above ($80\times 80$ grid, $\nEns=30$ ensemble members,
$n=300$ observations) and randomly permute the order of the observations before assimilation. 
\Cref{fig:synthetic_scores} compares the performance of sequential 
and non-sequential assimilation for $50$ different random orderings of the same dataset.

The experiments demonstrate that in a sequential setting, observation ordering can have an 
effect of roughly $5\%$ on the assimilation performance, thus confirming the results of \citep{nerger_ordering} 
and providing further proof that sequential assimilation is not consistent. As expected, all-at-once 
assimilation shows no dependence on data ordering.

\newpage
\subsubsection{Computational Capabilities and Limitations}
Even if all-at-once assimilation outperforms sequential assimilation in terms of reconstruction performance, as was shown in the 
previous section, there is still a computational overload associated to the computation of the full state covariance matrix. 
This results in a memory footprint that grows quadratically with the state dimension. In order to get a rough understanding 
of the limitations of all-at-once assimilation, it is worth studying the dependence of the memory footprint and total runtime 
on the dimension of the state vector.

\begin{figure}[h!]
    \centering
	\includegraphics[width=0.99\linewidth]{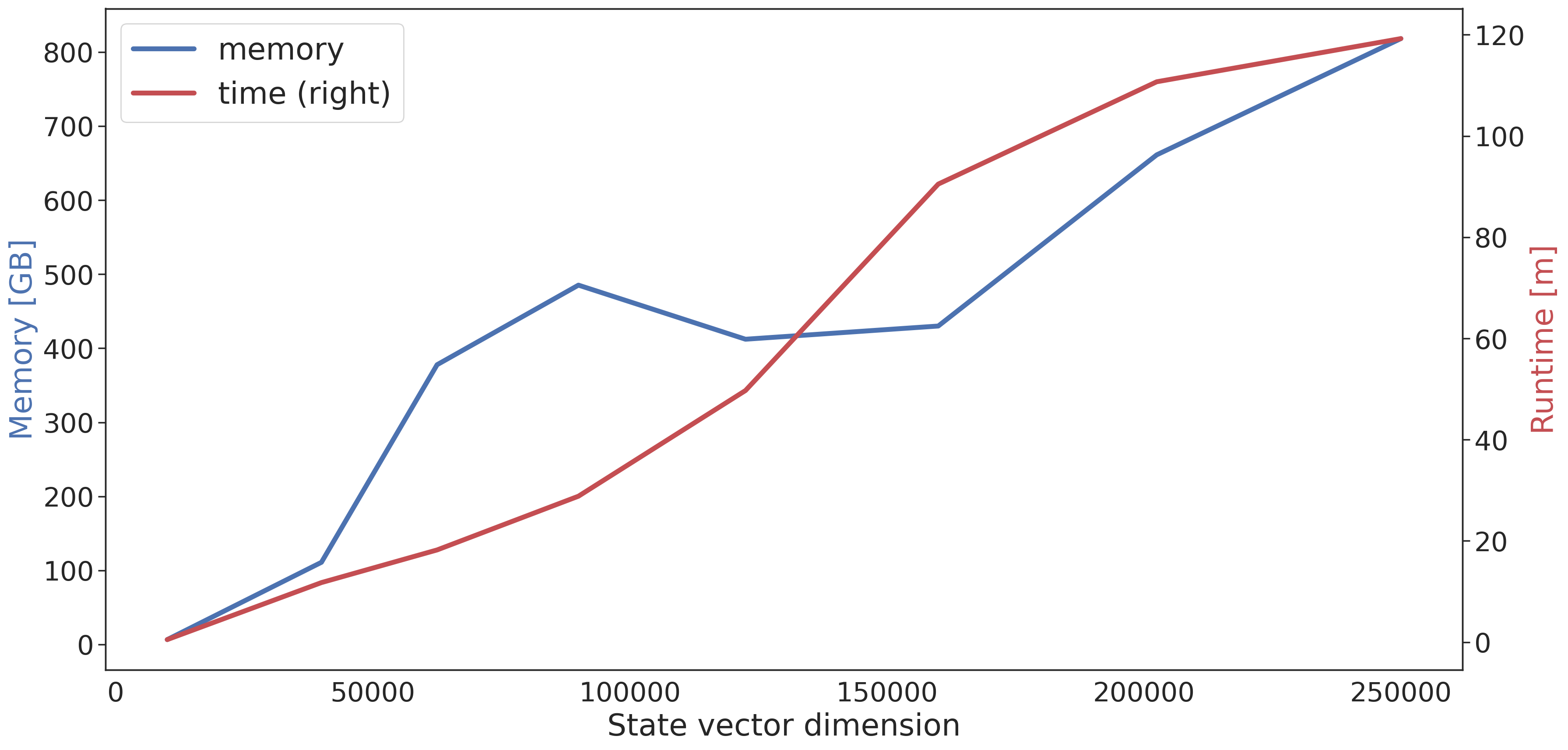}
    \caption{Peak memory usage and total runtime for all-at-once assimilation of 10000 data points as a function of the state dimension.}\label{fig:mem_usage}
\end{figure}

\Cref{fig:mem_usage} depicts the peak memory use and total runtime for assimilation of 10000 data points using our implementation 
of all at once assimilation for the EnSRF (see \Cref{sec:distributed}). The assimilation is run on a multi-node computing cluster 
managed by SLURM \citep{slurm} and the number of noder dedicated to the assimilation is allowed to grow if available memory becomes tight. 
For this reason, total memory usage and runtime depicted in \Cref{fig:mem_usage} should be considered more as rough estimates that showcase 
the capabilities of our implementation, since in practice, the total memory use will fluctuate depending on the hardware and specific 
settings of the cluster. The above results demonstrate that our framework is able to perform covariance localization 
and ensemble Kalman filtering in state spaces of dimension up to $250000$, which, in climate reconstruction applications, would 
correspond to performing reconstruction of several correlated climate variables on a high-resolution grid. Overall, those results 
indicate that our all-at-once assimilation framework is able to scale to realistic application settings, thus paving 
the way for the use of non-sequential assimilation in paleoclimate reconstruction tasks and other large-scale data assimilation problems.

\newpage
\subsection{Paleoclimate Reconstruction Test Case}\label{sec:paleoclimate_example}


In order to demonstrate the superiority of all-at-once EnSRF assimilation over the sequential 
one on real-world problems, we now turn to a paleoclimate reconstruction task and 
perform the same comparisons as in the last section. This application reproduces a 
popular Kalman-based method for climate and weather reconstruction \cite{bhend_ensemble_climate,twentieth_century_reanalysis,improved_twentieth_century,franke}. 
The idea is to blend historical observations with 
climate models simulations to get 
a global reconstruction of past climate states.

\begin{figure}[h]
    \centering
	\includegraphics[width=0.8\linewidth,height=0.28\linewidth]{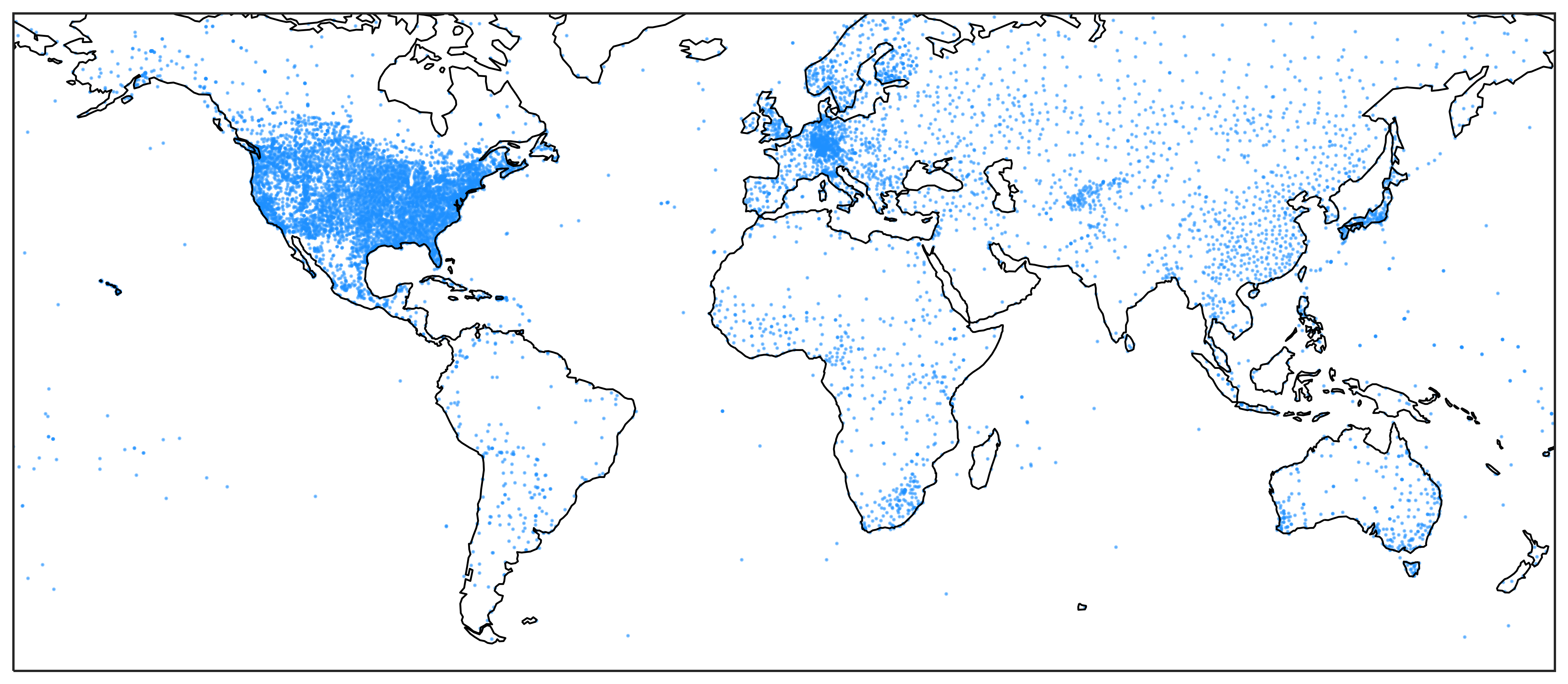}
    \caption{Stations in the ISTI dataset for January 1960.}
    \label{fig:stations_1960}
\end{figure}

The goal of our paleoclimatic reconstruction task is to reconstruct 
monthly average temperatures for the period 1960-1980 using an EnSRF. As a reference (groud truth) for the climate 
state over this period, we use the \textit{Climatic Research Unit gridded Time Series} (CRU TS) dataset \citep{Harris2020}. 
CRU TS is a high-resolution ($0.5^{\circ}$ by $0.5^{\circ}$) dataset of monthly reconstructed climate variables 
(mean temperature, precipitation rate, ...) produced by interpolating station data. For our practical purpose, 
this dataset can be considered as embodying the state of the art in climate reconstruction. 
Our starting ensemble is a set of $30$ model simulations from the general circulation model \textit{ECHAM5.4} \citep{roeckner1} discretized on a $192\time 96$ grid 
and is the same as used for several state-of-the-art climate reconstructions 
\citep{bhend_ensemble_climate,franke}. The simulations incorporate known external forcings such as 
solar irradiance, volcanic activity and greenhous gas concentration. To bring the simulations 
back towards reality, we assimilate station data of monthly average temperature. The dataset 
used for the station data is the \textit{international surface temperature initiative (ISTI) global land surface databank} stage 3 station dataset \citep{ilsd_dataset}. \Cref{fig:stations_1960} 
shows an overview of the stations available in the dataset for January 1960.

\begin{figure}[h]
    \centering
\begin{subfigure}{0.99\textwidth}
	\includegraphics[width=0.99\linewidth,height=0.27\linewidth]{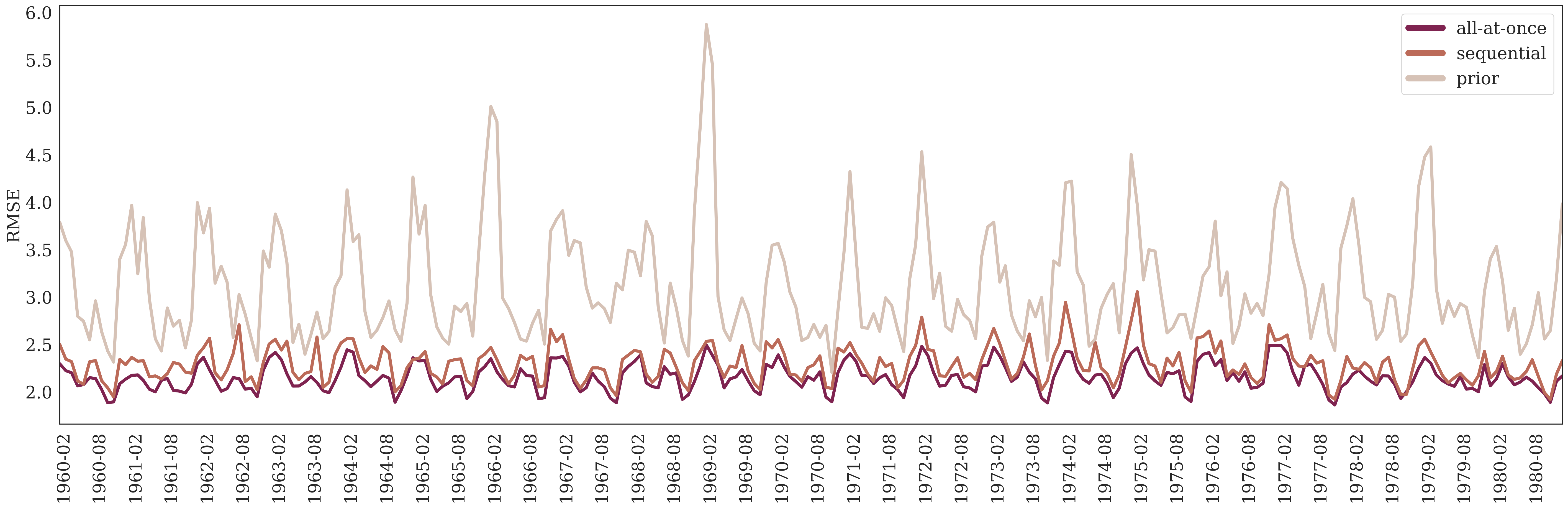}
\end{subfigure}\\
    \caption{RMSE of the reconstruction for different assimilation schemes.}
\label{fig:20th_rmse_comp}
\end{figure}

\begin{figure}[h]
    \centering
\begin{subfigure}{0.99\textwidth}
 	\includegraphics[width=0.99\linewidth,height=0.27\linewidth]{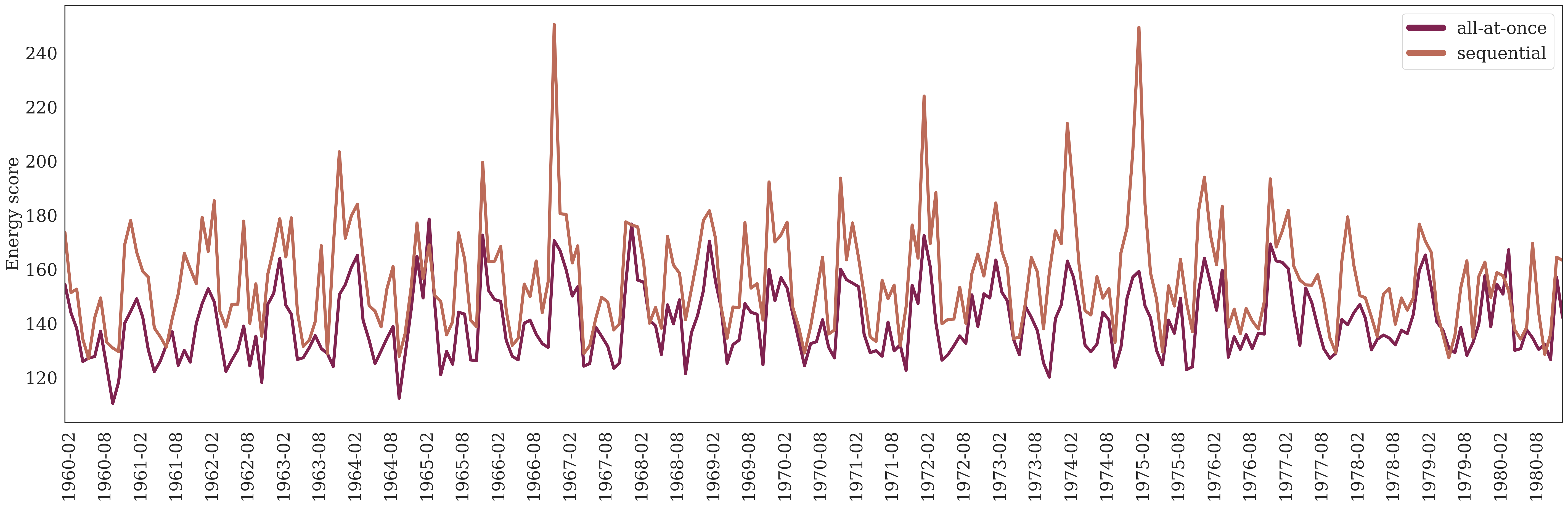}
 	\caption{Energy score}
 \end{subfigure}
    \caption{Energy score for different assimilation schemes.}
    \label{fig:20th_rmse_es_comp}
\end{figure}

As for the synthetic test case, we compare the RMSE skill score, energy score and RMSE 
for the different assimilation schemes. 
\Cref{fig:20th_rmse_comp,fig:20th_rmse_es_comp} show the monthly RMSE and ES and demonstrates how 
all-at-once assimilation outperforms sequential assimilation 
by a magnitude that is often of half a degree in the RMSE. Similarly, 
the energy score also indicates that all-at-once assimilation should be preferred 
over the sequential one.

\begin{figure}[h!]
    \centering
	\includegraphics[width=0.99\linewidth,height=0.25\linewidth]{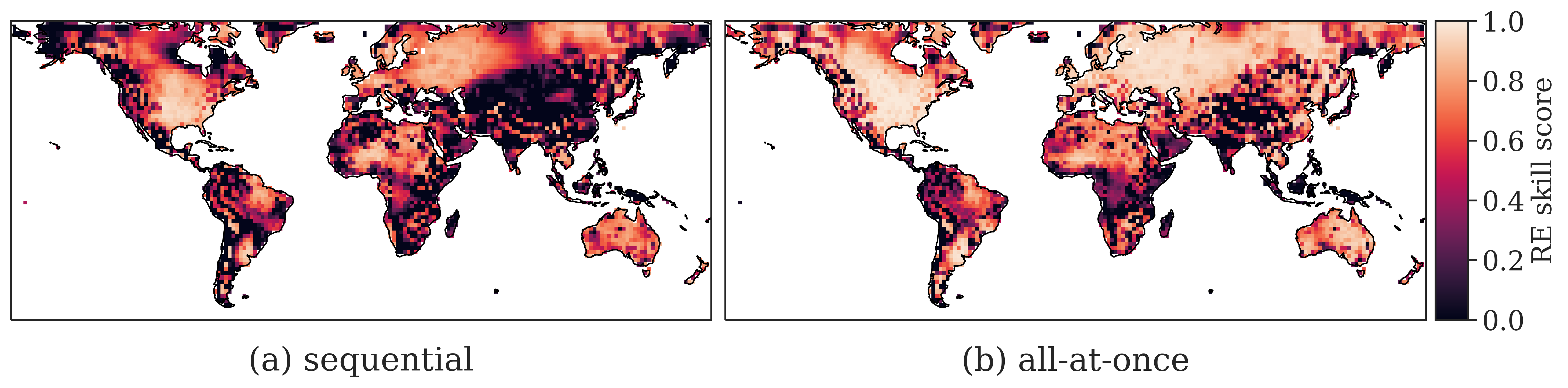}
\caption{RMSE skill scores for reconstruction of the reference dataset over the 1960-1980 period.}
\label{fig:20th_re_comp}
\end{figure}

Apart from the above metrics, one can also study the spatial distribution of the RMSE skill 
score aggregated over the whole assimilation window (\Cref{fig:20th_re_comp}). In this figure, 
negative values of the score (reconstruction poorer than background) are clamped at $0$. 
\newpage
\Cref{fig:20th_re_comp} is in line with the previous results, confirming 
that all-at-once outperforms sequential assimilation. We see that all-at-once assimilation is 
able to enhance the results of sequential assimilation in regions close to available 
observations (see \Cref{fig:stations_1960}), while performing poorly in regions where 
sequential assimilation also struggles. This seems to indicate that all-at-once assimilation
 is able to better handle spatial correlations.

\section{Conclusion}\label{seq:conclusion}
Leveraging modern distributed computing techniques provided 
by the \texttt{DASK} Python library, we are able to overcome the memory bottlenecks 
of Ensemble Kalman Filtering and provide a distributed implementation of 
the Ensemble Square Root Filter that allows for non-sequential data assimilation 
in problems of realistic sizes. We provide comparisons of our non-sequential, all-at-once 
assimilation scheme with the traditional sequential one on a synthetic assimilation 
problem and on a real-world paleoclimatic reconstruction task. Our benchmark 
shows that all-at-once assimilation outperforms sequential assimilation in 
both situations. In passing, we also provide an assessment of the depenced of 
sequential assimilation on observation ordering.

By allowing for non-sequential assimilation, our framework solves the long-standing 
problem of having to resort to inconsistent update equations in ensemble Kalman 
filtering and allows for a sytematic study of the effects of inconsistent updates 
on assimilation results.
Furthermore, we believe that distributed computing opens new venues of research in data 
assimilation and hope that our work can serve as a first paving stone in that direction. 
Among the next challenges that can be tackled using our distributed assimilation 
framework, we believe that the development of new covariance estimation techniques 
is the most promising one. Indeed, our implementation makes the full state covariance 
matrix available to the user, as well as a compressed representation of it through its 
truncated SVD, thereby allowing for the implementation of complex localization 
techniques in a large-scale setting.

%
%

\newpage
\section*{Open Research Section}
Climate simulations used for the starting ensemble in the paleoclimatic reconstruction example of \cref{sec:paleoclimate_example} 
is available online at \citep{ekf400_simulations}. Assimilated station data is available at \citep{ilsd_dataset}.

\subsection*{Acknowledgments}
C. T. acknowledges funding from the Swiss National Science Fundation under project nr. 178858. 
S.B., J.F. and C.T. acknowledge funding from the European Commission under Horizon 2020 (ERC grant 787574).
The authors would also like to thank Veronika Valler for participating 
in preliminary developments, Lorenz Hilfiker for helpful comments on a draft of this article 
and Alberto Terranova for testing the assimilation code.

%
%

\bibliographystyle{abbrvnat}
\bibliography{bibliography}

\begin{thebibliography}{49}
\providecommand{\natexlab}[1]{#1}
\providecommand{\url}[1]{\texttt{#1}}
\expandafter\ifx\csname urlstyle\endcsname\relax
  \providecommand{\doi}[1]{doi: #1}\else
  \providecommand{\doi}{doi: \begingroup \urlstyle{rm}\Url}\fi

\bibitem[Anderson and Anderson(1999)]{anderson_inflation}
J.~L. Anderson and S.~L. Anderson.
\newblock A monte carlo implementation of the nonlinear filtering problem to
  produce ensemble assimilations and forecasts.
\newblock \emph{Monthly Weather Review}, 127\penalty0 (12):\penalty0 2741 --
  2758, 1999.
\newblock \doi{10.1175/1520-0493(1999)127<2741:AMCIOT>2.0.CO;2}.
\newblock URL
  \url{https://journals.ametsoc.org/view/journals/mwre/127/12/1520-0493_1999_127_2741_amciot_2.0.co_2.xml}.

\bibitem[Bhend et~al.(2012)Bhend, Franke, Folini, Wild, and
  Br\"onnimann]{bhend_ensemble_climate}
J.~Bhend, J.~Franke, D.~Folini, M.~Wild, and S.~Br\"onnimann.
\newblock An ensemble-based approach to climate reconstructions.
\newblock \emph{Climate of the Past}, 8\penalty0 (3):\penalty0 963--976, 2012.
\newblock \doi{10.5194/cp-8-963-2012}.
\newblock URL \url{https://cp.copernicus.org/articles/8/963/2012/}.

\bibitem[Bjerregård et~al.(2021)Bjerregård, Møller, and
  Madsen]{bjerregard_multivariate_scoring}
M.~B. Bjerregård, J.~K. Møller, and H.~Madsen.
\newblock An introduction to multivariate probabilistic forecast evaluation.
\newblock \emph{Energy and AI}, 4:\penalty0 100058, 2021.
\newblock ISSN 2666-5468.
\newblock \doi{https://doi.org/10.1016/j.egyai.2021.100058}.
\newblock URL
  \url{https://www.sciencedirect.com/science/article/pii/S2666546821000124}.

\bibitem[Bopardikar(2017)]{bopardikar}
S.~D. Bopardikar.
\newblock Randomized matrix factorization for kalman filtering.
\newblock In \emph{2017 American Control Conference (ACC)}, pages 5795--5800,
  2017.
\newblock \doi{10.23919/ACC.2017.7963858}.

\bibitem[Burgers et~al.(1998)Burgers, van Leeuwen, and
  Evensen]{burgers_perturbed}
G.~Burgers, P.~J. van Leeuwen, and G.~Evensen.
\newblock Analysis scheme in the ensemble kalman filter.
\newblock \emph{Monthly Weather Review}, 126\penalty0 (6):\penalty0 1719 --
  1724, 1998.
\newblock \doi{10.1175/1520-0493(1998)126<1719:ASITEK>2.0.CO;2}.
\newblock URL
  \url{https://journals.ametsoc.org/view/journals/mwre/126/6/1520-0493_1998_126_1719_asitek_2.0.co_2.xml}.

\bibitem[Compo et~al.(2011)Compo, Whitaker, Sardeshmukh, Matsui, Allan, Yin,
  Gleason, Vose, Rutledge, Bessemoulin, Brönnimann, Brunet, Crouthamel, Grant,
  Groisman, Jones, Kruk, Kruger, Marshall, Maugeri, Mok, Nordli, Ross, Trigo,
  Wang, Woodruff, and Worley]{twentieth_century_reanalysis}
G.~P. Compo, J.~S. Whitaker, P.~D. Sardeshmukh, N.~Matsui, R.~J. Allan, X.~Yin,
  B.~E. Gleason, R.~S. Vose, G.~Rutledge, P.~Bessemoulin, S.~Brönnimann,
  M.~Brunet, R.~I. Crouthamel, A.~N. Grant, P.~Y. Groisman, P.~D. Jones, M.~C.
  Kruk, A.~C. Kruger, G.~J. Marshall, M.~Maugeri, H.~Y. Mok, {\O}.~Nordli,
  T.~F. Ross, R.~M. Trigo, X.~L. Wang, S.~D. Woodruff, and S.~J. Worley.
\newblock The twentieth century reanalysis project.
\newblock \emph{Quarterly Journal of the Royal Meteorological Society},
  137\penalty0 (654):\penalty0 1--28, 2011.
\newblock \doi{https://doi.org/10.1002/qj.776}.
\newblock URL
  \url{https://rmets.onlinelibrary.wiley.com/doi/abs/10.1002/qj.776}.

\bibitem[Cook et~al.(1994)Cook, Briffa, and Jones]{re_score}
E.~R. Cook, K.~R. Briffa, and P.~D. Jones.
\newblock Spatial regression methods in dendroclimatology: A review and
  comparison of two techniques.
\newblock \emph{International Journal of Climatology}, 14\penalty0
  (4):\penalty0 379--402, 1994.
\newblock \doi{https://doi.org/10.1002/joc.3370140404}.
\newblock URL
  \url{https://rmets.onlinelibrary.wiley.com/doi/abs/10.1002/joc.3370140404}.

\bibitem[{Dask Development Team}(2016)]{dask}
{Dask Development Team}.
\newblock \emph{Dask: Library for dynamic task scheduling}, 2016.
\newblock URL \url{https://dask.org}.

\bibitem[Efron and Morris(1976)]{efron_morris}
B.~Efron and C.~Morris.
\newblock Multivariate empirical bayes and estimation of covariance matrices.
\newblock \emph{The Annals of Statistics}, 4\penalty0 (1):\penalty0 22--32,
  1976.
\newblock ISSN 00905364.
\newblock URL \url{http://www.jstor.org/stable/2957992}.

\bibitem[Evensen(1994)]{evensen}
G.~Evensen.
\newblock Sequential data assimilation with a nonlinear quasi-geostrophic model
  using monte carlo methods to forecast error statistics.
\newblock \emph{Journal of Geophysical Research: Oceans}, 99\penalty0
  (C5):\penalty0 10143--10162, 1994.
\newblock \doi{https://doi.org/10.1029/94JC00572}.
\newblock URL
  \url{https://agupubs.onlinelibrary.wiley.com/doi/abs/10.1029/94JC00572}.

\bibitem[Evensen(2003)]{evensen_practical}
G.~Evensen.
\newblock The ensemble kalman filter: Theoretical formulation and practical
  implementation.
\newblock \emph{Ocean dynamics}, 53\penalty0 (4):\penalty0 343--367, 2003.

\bibitem[Evensen et~al.(2009)]{evensen_book}
G.~Evensen et~al.
\newblock \emph{Data assimilation: the ensemble Kalman filter}, volume~2.
\newblock Springer, 2009.

\bibitem[Farchi and Bocquet(2019)]{farchi_bocquet}
A.~Farchi and M.~Bocquet.
\newblock On the efficiency of covariance localisation of the ensemble kalman
  filter using augmented ensembles.
\newblock \emph{Frontiers in Applied Mathematics and Statistics}, 5, 02 2019.
\newblock \doi{10.3389/fams.2019.00003}.

\bibitem[Franke et~al.(2017)Franke, Br{\"o}nnimann, Bhend, and
  Brugnara]{franke}
J.~Franke, S.~Br{\"o}nnimann, J.~Bhend, and Y.~Brugnara.
\newblock A monthly global paleo-reanalysis of the atmosphere from 1600 to 2005
  for studying past climatic variations.
\newblock \emph{Scientific Data}, 4\penalty0 (1):\penalty0 170076, Jun 2017.
\newblock ISSN 2052-4463.
\newblock \doi{10.1038/sdata.2017.76}.
\newblock URL \url{https://doi.org/10.1038/sdata.2017.76}.

\bibitem[Furrer and Bengtsson(2007)]{furrer}
R.~Furrer and T.~Bengtsson.
\newblock Estimation of high-dimensional prior and posterior covariance
  matrices in kalman filter variants.
\newblock \emph{Journal of Multivariate Analysis}, 98\penalty0 (2):\penalty0
  227--255, 2007.
\newblock ISSN 0047-259X.
\newblock \doi{https://doi.org/10.1016/j.jmva.2006.08.003}.
\newblock URL
  \url{https://www.sciencedirect.com/science/article/pii/S0047259X06001187}.

\bibitem[Gelb et~al.(1974)]{gelb1974applied}
A.~Gelb et~al.
\newblock \emph{Applied optimal estimation}.
\newblock MIT press, 1974.

\bibitem[Gneiting and Raftery(2007)]{gneiting_scoring}
T.~Gneiting and A.~E. Raftery.
\newblock Strictly proper scoring rules, prediction, and estimation.
\newblock \emph{Journal of the American Statistical Association}, 102\penalty0
  (477):\penalty0 359--378, 2007.
\newblock \doi{10.1198/016214506000001437}.
\newblock URL \url{https://doi.org/10.1198/016214506000001437}.

\bibitem[Gneiting et~al.(2008)Gneiting, Stanberry, Grimit, Held, and
  Johnson]{gneiting_energy_score}
T.~Gneiting, L.~I. Stanberry, E.~P. Grimit, L.~Held, and N.~A. Johnson.
\newblock Assessing probabilistic forecasts of multivariate quantities, with an
  application to ensemble predictions of surface winds.
\newblock \emph{TEST}, 17\penalty0 (2):\penalty0 211--235, Aug 2008.
\newblock ISSN 1863-8260.
\newblock \doi{10.1007/s11749-008-0114-x}.
\newblock URL \url{https://doi.org/10.1007/s11749-008-0114-x}.

\bibitem[Grooms(2022)]{grooms2022comparison}
I.~Grooms.
\newblock A comparison of nonlinear extensions to the ensemble kalman filter:
  Gaussian anamorphosis and two-step ensemble filters.
\newblock \emph{Computational Geosciences}, 26\penalty0 (3):\penalty0 633--650,
  2022.

\bibitem[Haff(1980)]{haff}
L.~R. Haff.
\newblock {Empirical Bayes Estimation of the Multivariate Normal Covariance
  Matrix}.
\newblock \emph{The Annals of Statistics}, 8\penalty0 (3):\penalty0 586 -- 597,
  1980.
\newblock \doi{10.1214/aos/1176345010}.
\newblock URL \url{https://doi.org/10.1214/aos/1176345010}.

\bibitem[Halko et~al.(2011)Halko, Martinsson, and Tropp]{halko2011}
N.~Halko, P.-G. Martinsson, and J.~A. Tropp.
\newblock Finding structure with randomness: Probabilistic algorithms for
  constructing approximate matrix decompositions.
\newblock \emph{SIAM review}, 53\penalty0 (2):\penalty0 217--288, 2011.

\bibitem[Hamill et~al.(2001)Hamill, Whitaker, and Snyder]{hamill_localization}
T.~M. Hamill, J.~S. Whitaker, and C.~Snyder.
\newblock Distance-dependent filtering of background error covariance estimates
  in an ensemble kalman filter.
\newblock \emph{Monthly Weather Review}, 129\penalty0 (11):\penalty0 2776 --
  2790, 2001.
\newblock \doi{10.1175/1520-0493(2001)129<2776:DDFOBE>2.0.CO;2}.
\newblock URL
  \url{https://journals.ametsoc.org/view/journals/mwre/129/11/1520-0493_2001_129_2776_ddfobe_2.0.co_2.xml}.

\bibitem[Harris et~al.(2020)Harris, Osborn, Jones, and Lister]{Harris2020}
I.~Harris, T.~J. Osborn, P.~Jones, and D.~Lister.
\newblock Version 4 of the cru ts monthly high-resolution gridded multivariate
  climate dataset.
\newblock \emph{Scientific Data}, 7\penalty0 (1):\penalty0 109, Apr 2020.
\newblock ISSN 2052-4463.
\newblock \doi{10.1038/s41597-020-0453-3}.
\newblock URL \url{https://doi.org/10.1038/s41597-020-0453-3}.

\bibitem[Houtekamer and Mitchell(2001)]{houtekamer_sequential}
P.~L. Houtekamer and H.~L. Mitchell.
\newblock A sequential ensemble kalman filter for atmospheric data
  assimilation.
\newblock \emph{Monthly Weather Review}, 129\penalty0 (1):\penalty0 123 -- 137,
  2001.
\newblock \doi{10.1175/1520-0493(2001)129<0123:ASEKFF>2.0.CO;2}.
\newblock URL
  \url{https://journals.ametsoc.org/view/journals/mwre/129/1/1520-0493_2001_129_0123_asekff_2.0.co_2.xml}.

\bibitem[Houtekamer and Zhang(2016)]{houtemaker_review}
P.~L. Houtekamer and F.~Zhang.
\newblock Review of the ensemble kalman filter for atmospheric data
  assimilation.
\newblock \emph{Monthly Weather Review}, 144\penalty0 (12):\penalty0 4489 --
  4532, 2016.
\newblock \doi{https://doi.org/10.1175/MWR-D-15-0440.1}.
\newblock URL
  \url{https://journals.ametsoc.org/view/journals/mwre/144/12/mwr-d-15-0440.1.xml}.

\bibitem[Jordan et~al.(2019)Jordan, Krüger, and Lerch]{jordan_scoring}
A.~Jordan, F.~Krüger, and S.~Lerch.
\newblock Evaluating probabilistic forecasts with scoringrules.
\newblock \emph{Journal of Statistical Software}, 90\penalty0 (12):\penalty0
  1–37, 2019.
\newblock \doi{10.18637/jss.v090.i12}.
\newblock URL
  \url{https://www.jstatsoft.org/index.php/jss/article/view/v090i12}.

\bibitem[Julier and Uhlmann(1997)]{julier1997new}
S.~J. Julier and J.~K. Uhlmann.
\newblock New extension of the kalman filter to nonlinear systems.
\newblock In \emph{Signal processing, sensor fusion, and target recognition
  VI}, volume 3068, pages 182--193. Spie, 1997.

\bibitem[Katzfuss et~al.(2016)Katzfuss, Stroud, and Wikle]{katzfuss_enkf}
M.~Katzfuss, J.~R. Stroud, and C.~K. Wikle.
\newblock Understanding the ensemble kalman filter.
\newblock \emph{The American Statistician}, 70\penalty0 (4):\penalty0 350--357,
  2016.
\newblock \doi{10.1080/00031305.2016.1141709}.
\newblock URL \url{https://doi.org/10.1080/00031305.2016.1141709}.

\bibitem[Ledoit and Wolf(2004)]{ledoit_wolf}
O.~Ledoit and M.~Wolf.
\newblock A well-conditioned estimator for large-dimensional covariance
  matrices.
\newblock \emph{Journal of Multivariate Analysis}, 88\penalty0 (2):\penalty0
  365--411, 2004.
\newblock ISSN 0047-259X.
\newblock \doi{https://doi.org/10.1016/S0047-259X(03)00096-4}.
\newblock URL
  \url{https://www.sciencedirect.com/science/article/pii/S0047259X03000964}.

\bibitem[Ledoit and Wolf(2012)]{ledoit_wolf_nonlinear}
O.~Ledoit and M.~Wolf.
\newblock {Nonlinear shrinkage estimation of large-dimensional covariance
  matrices}.
\newblock \emph{The Annals of Statistics}, 40\penalty0 (2):\penalty0 1024 --
  1060, 2012.
\newblock \doi{10.1214/12-AOS989}.
\newblock URL \url{https://doi.org/10.1214/12-AOS989}.

\bibitem[Lefebvre et~al.(2004)Lefebvre, Bruyninckx, and
  De~Schutter]{lefebvre2004kalman}
T.~Lefebvre, H.~Bruyninckx, and J.~De~Schutter.
\newblock Kalman filters for non-linear systems: a comparison of performance.
\newblock \emph{International journal of Control}, 77\penalty0 (7):\penalty0
  639--653, 2004.

\bibitem[{Max Planck Institute for Meteorology
  (MPI-M)}(2017)]{ekf400_simulations}
{Max Planck Institute for Meteorology (MPI-M)}.
\newblock Input for ekf400: Original echam simulations (ccc400) and assimilated
  observations, 2017.
\newblock URL
  \url{http://hdl.handle.net/21.14106/0bb12ac5cde477d3f382d9440e29d718e0943257}.

\bibitem[Murphy and Epstein(1989)]{re_score_murphy}
A.~H. Murphy and E.~S. Epstein.
\newblock Skill scores and correlation coefficients in model verification.
\newblock \emph{Monthly Weather Review}, 117\penalty0 (3):\penalty0 572 -- 582,
  1989.
\newblock
  \doi{https://doi.org/10.1175/1520-0493(1989)117<0572:SSACCI>2.0.CO;2}.
\newblock URL
  \url{https://journals.ametsoc.org/view/journals/mwre/117/3/1520-0493_1989_117_0572_ssacci_2_0_co_2.xml}.

\bibitem[Nerger(2015)]{nerger_ordering}
L.~Nerger.
\newblock On serial observation processing in localized ensemble kalman
  filters.
\newblock \emph{Monthly Weather Review}, 143\penalty0 (5):\penalty0 1554 --
  1567, 2015.
\newblock \doi{10.1175/MWR-D-14-00182.1}.
\newblock URL
  \url{https://journals.ametsoc.org/view/journals/mwre/143/5/mwr-d-14-00182.1.xml}.

\bibitem[Rasmussen and Williams(2006)]{rasmussen_williams}
C.~E. Rasmussen and C.~K.~I. Williams.
\newblock \emph{{G}aussian Processes for Machine Learning}.
\newblock Adaptive computation and machine learning. MIT Press, 2006.

\bibitem[Rennie et~al.(2014)Rennie, Lawrimore, Gleason, Thorne, Morice, Menne,
  Williams, de~Almeida, Christy, Flannery, Ishihara, Kamiguchi, Klein-Tank,
  Mhanda, Lister, Razuvaev, Renom, Rusticucci, Tandy, Worley, Venema, Angel,
  Brunet, Dattore, Diamond, Lazzara, Le~Blancq, Luterbacher, Mächel,
  Revadekar, Vose, and Yin]{ilsd_dataset}
J.~J. Rennie, J.~H. Lawrimore, B.~E. Gleason, P.~W. Thorne, C.~P. Morice, M.~J.
  Menne, C.~N. Williams, W.~G. de~Almeida, J.~Christy, M.~Flannery,
  M.~Ishihara, K.~Kamiguchi, A.~M.~G. Klein-Tank, A.~Mhanda, D.~H. Lister,
  V.~Razuvaev, M.~Renom, M.~Rusticucci, J.~Tandy, S.~J. Worley, V.~Venema,
  W.~Angel, M.~Brunet, B.~Dattore, H.~Diamond, M.~A. Lazzara, F.~Le~Blancq,
  J.~Luterbacher, H.~Mächel, J.~Revadekar, R.~S. Vose, and X.~Yin.
\newblock The international surface temperature initiative global land surface
  databank: monthly temperature data release description and methods.
\newblock \emph{Geoscience Data Journal}, 1\penalty0 (2):\penalty0 75--102,
  2014.
\newblock \doi{https://doi.org/10.1002/gdj3.8}.
\newblock URL
  \url{https://rmets.onlinelibrary.wiley.com/doi/abs/10.1002/gdj3.8}.

\bibitem[Rocklin(2015)]{dask_review}
M.~Rocklin.
\newblock Dask: Parallel computation with blocked algorithms and task
  scheduling.
\newblock In K.~Huff and J.~Bergstra, editors, \emph{Proceedings of the 14th
  Python in Science Conference}, pages 130 -- 136, 2015.

\bibitem[Roeckner et~al.(2003)Roeckner, Bäuml, Bonaventura, Brokopf, Esch,
  Giorgetta, Hagemann, Kirchner, Kornblueh, Manzini, Rhodin, Schlese,
  Schulzweida, and Tompkins]{roeckner1}
E.~Roeckner, G.~Bäuml, L.~Bonaventura, R.~Brokopf, M.~Esch, M.~Giorgetta,
  S.~Hagemann, I.~Kirchner, L.~Kornblueh, E.~Manzini, A.~Rhodin, U.~Schlese,
  U.~Schulzweida, and A.~Tompkins.
\newblock The atmospheric general circulation model echam 5. part i: model
  description.
\newblock \emph{Report / Max Planck Institute for Meteorology}, 349, 01 2003.

\bibitem[Slivinski et~al.(2019)Slivinski, Compo, Whitaker, Sardeshmukh, Giese,
  McColl, Allan, Yin, Vose, Titchner, Kennedy, Spencer, Ashcroft, Br\"onnimann,
  Brunet, Camuffo, Cornes, Cram, Crouthamel, Dom\'inguez-Castro, Freeman,
  Gergis, Hawkins, Jones, Jourdain, Kaplan, Kubota, Blancq, Lee, Lorrey,
  Luterbacher, Maugeri, Mock, Moore, Przybylak, Pudmenzky, Reason, Slonosky,
  Smith, Tinz, Trewin, Valente, Wang, Wilkinson, Wood, and
  Wyszy\`{i}nski]{improved_twentieth_century}
L.~C. Slivinski, G.~P. Compo, J.~S. Whitaker, P.~D. Sardeshmukh, B.~S. Giese,
  C.~McColl, R.~Allan, X.~Yin, R.~Vose, H.~Titchner, J.~Kennedy, L.~J. Spencer,
  L.~Ashcroft, S.~Br\"onnimann, M.~Brunet, D.~Camuffo, R.~Cornes, T.~A. Cram,
  R.~Crouthamel, F.~Dom\'inguez-Castro, J.~E. Freeman, J.~Gergis, E.~Hawkins,
  P.~D. Jones, S.~Jourdain, A.~Kaplan, H.~Kubota, F.~L. Blancq, T.-C. Lee,
  A.~Lorrey, J.~Luterbacher, M.~Maugeri, C.~J. Mock, G.~K. Moore, R.~Przybylak,
  C.~Pudmenzky, C.~Reason, V.~C. Slonosky, C.~A. Smith, B.~Tinz, B.~Trewin,
  M.~A. Valente, X.~L. Wang, C.~Wilkinson, K.~Wood, and P.~Wyszy\`{i}nski.
\newblock Towards a more reliable historical reanalysis: Improvements for
  version 3 of the twentieth century reanalysis system.
\newblock \emph{Quarterly Journal of the Royal Meteorological Society},
  145\penalty0 (724):\penalty0 2876--2908, 2019.
\newblock \doi{https://doi.org/10.1002/qj.3598}.
\newblock URL
  \url{https://rmets.onlinelibrary.wiley.com/doi/abs/10.1002/qj.3598}.

\bibitem[Snyder(2014)]{snyder_intro_kalman}
C.~Snyder.
\newblock {Introduction to the Kalman filter}.
\newblock In \emph{{Advanced Data Assimilation for Geosciences: Lecture Notes
  of the Les Houches School of Physics: Special Issue, June 2012}}. Oxford
  University Press, 10 2014.
\newblock ISBN 9780198723844.
\newblock \doi{10.1093/acprof:oso/9780198723844.003.0003}.
\newblock URL \url{https://doi.org/10.1093/acprof:oso/9780198723844.003.0003}.

\bibitem[Tippett et~al.(2003)Tippett, Anderson, Bishop, Hamill, and
  Whitaker]{tippett_square_root}
M.~K. Tippett, J.~L. Anderson, C.~H. Bishop, T.~M. Hamill, and J.~S. Whitaker.
\newblock Ensemble square root filters.
\newblock \emph{Monthly weather review}, 131\penalty0 (7):\penalty0 1485--1490,
  2003.

\bibitem[Travelletti et~al.(2023)Travelletti, Ginsbourger, and
  Linde]{travelletti2021}
C.~Travelletti, D.~Ginsbourger, and N.~Linde.
\newblock Uncertainty quantification and experimental design for large-scale
  linear inverse problems under gaussian process priors.
\newblock \emph{SIAM/ASA Journal on Uncertainty Quantification}, 11\penalty0
  (1):\penalty0 168--198, 2023.
\newblock \doi{10.1137/21M1445028}.
\newblock URL \url{https://doi.org/10.1137/21M1445028}.

\bibitem[Valler et~al.(2019)Valler, Franke, and Br\"onnimann]{valler}
V.~Valler, J.~Franke, and S.~Br\"onnimann.
\newblock Impact of different estimations of the background-error covariance
  matrix on climate reconstructions based on data assimilation.
\newblock \emph{Climate of the Past}, 15\penalty0 (4):\penalty0 1427--1441,
  2019.
\newblock \doi{10.5194/cp-15-1427-2019}.
\newblock URL \url{https://cp.copernicus.org/articles/15/1427/2019/}.

\bibitem[Wan and Van Der~Merwe(2000)]{wan2000unscented}
E.~A. Wan and R.~Van Der~Merwe.
\newblock The unscented kalman filter for nonlinear estimation.
\newblock In \emph{Proceedings of the IEEE 2000 Adaptive Systems for Signal
  Processing, Communications, and Control Symposium (Cat. No. 00EX373)}, pages
  153--158. Ieee, 2000.

\bibitem[Welch and Bishop(2000)]{welch_intro_kalman}
G.~Welch and G.~Bishop.
\newblock An introduction to the kalman filter.
\newblock Technical Report TR95041, Department of Computer Science, University
  of North Carolina, Chapel Hill, 2000.

\bibitem[Wheatcroft(2019)]{wheatcroft_skill_score}
E.~Wheatcroft.
\newblock Interpreting the skill score form of forecast performance metrics.
\newblock \emph{International Journal of Forecasting}, 35\penalty0
  (2):\penalty0 573--579, 2019.
\newblock ISSN 0169-2070.
\newblock \doi{https://doi.org/10.1016/j.ijforecast.2018.11.010}.
\newblock URL
  \url{https://www.sciencedirect.com/science/article/pii/S0169207019300093}.

\bibitem[Whitaker and Hamill(2002)]{whitaker_no_perturb}
J.~S. Whitaker and T.~M. Hamill.
\newblock Ensemble data assimilation without perturbed observations.
\newblock \emph{Monthly Weather Review}, 130\penalty0 (7):\penalty0 1913 --
  1924, 2002.
\newblock \doi{10.1175/1520-0493(2002)130<1913:EDAWPO>2.0.CO;2}.
\newblock URL
  \url{https://journals.ametsoc.org/view/journals/mwre/130/7/1520-0493_2002_130_1913_edawpo_2.0.co_2.xml}.

\bibitem[Whitaker et~al.(2008)Whitaker, Hamill, Wei, Song, and
  Toth]{whitaker_ordering}
J.~S. Whitaker, T.~M. Hamill, X.~Wei, Y.~Song, and Z.~Toth.
\newblock Ensemble data assimilation with the ncep global forecast system.
\newblock \emph{Monthly Weather Review}, 136\penalty0 (2):\penalty0 463--482,
  2008.

\bibitem[Yoo et~al.(2003)Yoo, Jette, and Grondona]{slurm}
A.~B. Yoo, M.~A. Jette, and M.~Grondona.
\newblock Slurm: Simple linux utility for resource management.
\newblock In D.~Feitelson, L.~Rudolph, and U.~Schwiegelshohn, editors,
  \emph{Job Scheduling Strategies for Parallel Processing}, pages 44--60,
  Berlin, Heidelberg, 2003. Springer Berlin Heidelberg.
\newblock ISBN 978-3-540-39727-4.

\end{thebibliography}

%
%
%
%
%

\end{document}